\documentclass{new_tlp}
\usepackage[utf8]{inputenc}
\usepackage{mathptmx}
\usepackage{amsmath}
\usepackage{amssymb}
\usepackage{xspace}
\usepackage{multirow}
\usepackage{calc}
\usepackage{url}

\usepackage{textcomp}
\usepackage{listings}
\makeatletter
\lst@Key{countblanklines}{true}[t]{\lstKV@SetIf{#1}\lst@ifcountblanklines}
\lst@AddToHook{OnEmptyLine}{%
    \lst@ifnumberblanklines\else%
    \lst@ifcountblanklines\else%
    \advance\c@lstnumber-\@ne\relax%
    \fi%
    \fi}
\makeatother
\lstdefinelanguage{asp}{
    breakatwhitespace=true,
    captionpos=b,
    numbers=left,
    numbersep=5pt,
    numberblanklines=false,
    countblanklines=false,
    commentstyle=\color{gray},
    frame=bt, framexbottommargin=5pt, framextopmargin=5pt,
    aboveskip=5pt, belowskip=5pt,
    abovecaptionskip=10pt
}
\lstnewenvironment{asp}{
	\lstset{
		language=asp,
		showstringspaces=false,
        keepspaces=true,
		formfeed=\newpage,
		tabsize=4,
		numbers=none,
		breaklines=true,
		literate={~} {$\sim$}{1},
    frame=none,
	}
}{}

\usepackage{tikz}\usetikzlibrary{arrows,fit,shapes.symbols,patterns}
\usepackage{caption}
\usepackage{subcaption}
\newtheorem{examplee}{Example}
\newenvironment{example}{\begin{examplee}}{\hfill $\blacksquare$\end{examplee}}

\newenvironment{rev}{\color{blue}}{\color{black}}

\submitted{n/a}
\revised{n/a}
\accepted{n/a}

\title[Aggregate Semantics for Propositional Answer Set Programs]{Aggregate Semantics for Propositional Answer Set Programs}
\author[M.~Alviano, W.~Faber and M.~Gebser]{%
  Mario Alviano\\ 
  University of Calabria\\
  \email{alviano@mat.unical.it}
  \and
  Wolfgang Faber and
  Martin Gebser\thanks{also affiliated with the Graz University of Technology}\\
  University of Klagenfurt\\
  \email{\{wolfgang.faber,martin.gebser\}@aau.at}
}

\newcommand{\mathtext}[1]{\ensuremath{\textsc{#1}}}
\newcommand{\at}{\ensuremath{\mathcal{P}}\xspace}
\newcommand{\agg}{\mathtext{agg}\xspace}
\newcommand{\op}{\ensuremath{\odot}\xspace}
\newcommand{\bound}{\ensuremath{w_0}\xspace}
\newcommand{\lits}[1]{\ensuremath{[#1]}}
\newcommand{\asum}{\mathtext{sum}\xspace}
\newcommand{\amin}{\mathtext{min}\xspace}
\newcommand{\amax}{\mathtext{max}\xspace}
\newcommand{\aavg}{\mathtext{avg}\xspace}
\newcommand{\atimes}{\mathtext{times}\xspace}
\newcommand{\acount}{\mathtext{count}\xspace}
\newcommand{\aeven}{\mathtext{even}\xspace}
\newcommand{\aodd}{\mathtext{odd}\xspace}
\newcommand{\atom}{\mathtext{at}\xspace}
\newcommand{\atoms}[1]{\ensuremath{\atom(#1)}}

\newcommand{\head}[1]{\ensuremath{H(#1)}}
\newcommand{\body}[1]{\ensuremath{B(#1)}}
\newcommand{\prg}{\ensuremath{P}\xspace}
\newcommand{\multi}[2]{\ensuremath{\omega(#1,#2)}}
\newcommand{\eval}[2]{\ensuremath{\alpha(#1,#2)}}
\newcommand{\undef}{\ensuremath{\varepsilon}\xspace}
\newcommand{\nmodels}{\ensuremath{\not\models}\xspace}

\newcommand{\lb}[1]{\ensuremath{\mathtext{lb}(#1)}}
\newcommand{\ub}[1]{\ensuremath{\mathtext{ub}(#1)}}

\newcommand{\ifaoif}{if and only if\xspace}
\newcommand{\poly}{\mathtext{P}\xspace}
\newcommand{\NP}{\mathtext{NP}\xspace}
\newcommand{\coNP}{\mathtext{coNP}\xspace}
\newcommand{\stwo}{\ensuremath{\Sigma_2^\poly}\xspace}
\newcommand{\checkas}{\textbf{Check}\xspace}
\newcommand{\existas}{\textbf{Exist}\xspace}

\newcommand{\reduct}[2]{\ensuremath{#1^{#2}}}
\newcommand{\fflp}{\mathtext{FFLP}\xspace}
\newcommand{\gz}{\mathtext{GZ}\xspace}
\newcommand{\lpst}{\mathtext{LPST}\xspace}
\newcommand{\mr}{\mathtext{MR}\xspace}
\newcommand{\pdb}{\mathtext{DPB}\xspace}
\newcommand{\any}{\ensuremath{\Delta}\xspace}

\newcommand{\tpo}{\ensuremath{\mathcal{T}}\xspace}
\newcommand{\tp}[1]{\ensuremath{\tpo_{#1}}}
\newcommand{\sat}[1]{\ensuremath{\models_{#1}}}
\newcommand{\unsat}[1]{\ensuremath{\nmodels_{#1}}}
\newcommand{\tpx}[3]{\ensuremath{\tp{#2,#3}^{#1}}}
\newcommand{\tpc}[4]{\ensuremath{\tpx{#1}{#2}{#3}(#4)}}
\newcommand{\tpu}[4]{\ensuremath{\tpx{#1}{#2}{#3}\uparrow #4}}

\newcommand{\gpo}{\ensuremath{\mathcal{G}}\xspace}
\newcommand{\gp}[1]{\ensuremath{\gpo_{#1}}}

\newcommand{\nap}{\ensuremath{\mathtext{not}}\xspace}
\newcommand{\naf}[1]{\ensuremath{\nap\ #1}}
\newcommand{\lit}{\ensuremath{\ell}\xspace}
\newcommand{\opp}[1]{\ensuremath{\overline{#1}}}
\newcommand{\nwf}{\ensuremath{\tau}\xspace}
\newcommand{\nwt}[1]{\nwf(#1)}

\newcommand{\aspcore}{\emph{ASP-Core-2}\xspace}

\newcommand{\capvspace}{\vspace{1em}}
\newcommand{\pacvspace}{\vspace{-0.8em}}
\newcommand{\tabhspace}{\hspace{0.5em}}
\newcommand{\tabrspace}{\renewcommand{\arraystretch}{1.2}}
\newlength{\tabespace}
\newcommand{\tabbspace}{\hspace*{-\tabespace}}

\begin{document}

\maketitle

\begin{abstract}
Answer Set Programming (ASP) emerged in the late 1990ies as a paradigm
for Knowledge Representation and Reasoning. 
The attractiveness of ASP builds on an expressive high-level modeling language
along with the availability of powerful off-the-shelf solving systems.
While the utility of incorporating aggregate expressions in the modeling language
has been realized almost simultaneously with the inception of the first ASP solving systems,
a general semantics of aggregates and its efficient implementation have been
long-standing challenges.
Aggregates have been proposed and widely used in database systems, and also in the deductive database language Datalog, which is one of the main precursors of ASP.
The use of
aggregates was, however, still restricted in Datalog (by either disallowing recursion or only allowing monotone aggregates), while several ways to integrate unrestricted aggregates evolved
in the context of ASP.
In this survey, we pick up at this point of development by presenting and comparing the main aggregate semantics that have been proposed for propositional ASP programs.
We highlight crucial properties such as computational complexity and expressive power,
and outline the capabilities and limitations of different approaches by 
illustrative examples.

\textbf{Under consideration in Theory and Practice of Logic Programming (TPLP)}
\end{abstract}

\begin{keywords}
answer set programming, aggregate expressions, semantics, complexity and expressiveness
\end{keywords}

\section{Introduction}\label{sec:introduction}

Answer Set Programming (ASP) \cite{breitr11a,gelleo02a,lifschitz02a,martru99a,niemela99a} is a paradigm for Knowledge Representation and Reasoning. 
ASP knowledge bases are encoded by means of logic rules interpreted according to the stable model semantics \cite{gellif88b,gellif91a}, that is, models of ASP programs are required to satisfy an additional stability condition guaranteeing that all true atoms in a model are necessary.
A strength of ASP is its high-level modeling language, capable of expressing all problems in the complexity classes \NP and \stwo by means of declarative statements \cite{daeigovo01a,schlipf95a}\begin{rev}, depending on the availability of disjunctive rules or unrestricted aggregates\end{rev}.
The attractiveness of ASP also builds on the availability of powerful off-the-shelf solving systems, among them \textsc{clingo} \cite{gekakasc17a}, \textsc{dlv} \cite{alcadofuleperiveza17a}, and \textsc{idp} \cite{brblbocapojalaradeve15a}.

The language of ASP offers several constructs to ease the representation of practical knowledge.
Aggregate expressions received particular interest by ASP designers  \begin{rev}\cite{baleme11a,depebr01a,fapfle11a,ferraris11a,gehakalisc15a,gelzha19a,liposotr10a,marrem04a,pedebr07a,siniso02a}\end{rev}, and the utility of their incorporation in the modeling language has been realized almost simultaneously with the inception of the first ASP solving systems.
In fact, aggregate expressions provide a natural syntax for expressing properties on sets of atoms collectively, which is often desired when modeling complex knowledge.
For example, aggregate expressions are widely used to enforce collective conditions on guessed relations. In \emph{graph k-colorability} \cite{garjoh79}, the guessed assignment of colors must be a total function, which can be enforced by means of the following constraint:
\begin{asp}
  :- vertex(X), #count{C : assign_color(X,C)} != 1.
\end{asp}
In \emph{independent set} \cite{garjoh79}, the guessed set of nodes must not be smaller than a given bound $k \geq 1$, which can be enforced by the following constraint:
\begin{asp}
  :- #count{X : in_independent_set(X)} < k.
\end{asp}

\begin{figure}
    \figrule
    \begin{subfigure}{.35\textwidth}
        \centering

        \tikzstyle{none} = [text centered]
        \tikzstyle{line} = [draw]
           
        \begin{tikzpicture}
            \node at (0,0) [none](a) {a};
            \node at (-1,-1) [none](b) {b};
            \node at (1,-1) [none](c) {c};
            \node at (-1.5,-2) [none](d) {d};
            \node at (-0.5,-2) [none](e) {e};
            \node at (0.25,-2) [none](f) {f};
            \node at (1,-2) [none](g) {g};
            \node at (1.75,-2) [none](h) {h};
            
            \path[line] (a) -- (b);
            \path[line] (a) -- (c);
            \path[line] (b) -- (d);
            \path[line] (b) -- (e);
            \path[line] (c) -- (f);
            \path[line] (c) -- (g);
            \path[line] (c) -- (h);
        \end{tikzpicture}
    \end{subfigure}
    \begin{subfigure}{.5\textwidth}
        \begin{asp}
tree(a).
tree(b). tree(c). 
tree(d). tree(e).
tree(f). tree(g). tree(h).

child(a,b). child(a,c).
child(b,d). child(b,e).
child(c,f). child(c,g). child(c,h).
        \end{asp}
    \end{subfigure}
    \begin{subfigure}{.125\textwidth}
        \begin{asp}
nodes(a,8).
nodes(b,3).
nodes(c,4).
nodes(d,1).
nodes(e,1).
nodes(f,1).
nodes(g,1).
nodes(h,1).
        \end{asp}
    \end{subfigure}
    \pacvspace
    \caption{A tree, its relational representation, and the number of nodes in each (sub)tree.}\label{fig:tree}
    \figrule
\end{figure}
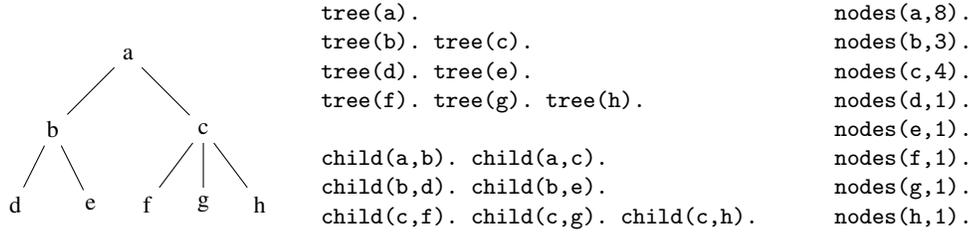

Aggregate expressions in constraints, as in the examples above, are frequent and leave no ambiguity in their intended semantics.
However, there are other frequent use cases in which aggregation functions are involved in inductive definitions, which open the possibility of  using aggregation functions for reasoning about recursive data structures.
For example, the number of nodes in a tree (and its subtrees) can be determined by the following rule:
\begin{asp}
  nodes(T,C+1) :- tree(T), C = #sum{C',T' : child(T,T'), nodes(T',C')}.
\end{asp}
Figure~\ref{fig:tree} shows a tree and the expected outcome for the \lstinline|nodes| relation.
Note that the sum is applied to a multiset, specifically to the multiset obtained by projecting on the first element the following set of tuples:
\lstinline|{(C',T') ${}\mid{}$ child(T,T') ${}\wedge{}$ nodes(T',C')}|.
For example, given \lstinline|nodes(d,1)| and \lstinline|nodes(e,1)|, for $\mathtt{T = b}$ the set is \lstinline|{(1,d), (1,e)}|, and therefore the sum aggregation is applied to the multiset $[1, 1]$.

\begin{figure}
    \figrule
    \begin{subfigure}{.3\textwidth}
        \centering

        \tikzstyle{none} = [text centered]
        \tikzstyle{line} = [draw, ->]
           
        \begin{tikzpicture}
            \node at (0,0) [none](a) {a};
            \node at (-1,-1) [none](b) {b};
            \node at (1,-1) [none](c) {c};
            
            \path[line] (a) -- node [midway, left] {80\%} (b);
            \path[line] (a) -- node [midway, right] {30\%} (c);
            \path[line] (b) -- node [midway, above] {30\%} (c);
        \end{tikzpicture}
    \end{subfigure}
    \begin{subfigure}{.35\textwidth}
        \begin{asp}
company(a). owns(a,b,80).
company(b). owns(a,c,30).
company(c). owns(b,c,30).
        \end{asp}
    \end{subfigure}
    \begin{subfigure}{.25\textwidth}
        \begin{asp}
controls(a,b).
controls(a,c).
        \end{asp}
    \end{subfigure}
    \caption{Companies and their shares, their relational representation, and the control relationships.}\label{fig:company}
    \figrule
\end{figure}
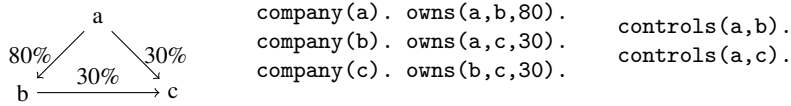

As another example, consider the \emph{company controls} problem, originally proposed \begin{rev}by \citeN{mupira90a}\end{rev}, where a set of companies and their stock shares are given and the task is to determine the control relationships between companies.
The following rule encodes such   knowledge:
\begin{asp}
  controls(A,B) :- company(A), company(B),
      #sum{S : own(A,B,S); S,C : controls(A,C), own(C,B,S), A != C} > 50.
\end{asp}
%
Figure~\ref{fig:company} shows an instance of this problem and the expected outcome.
Non-deterministic variants of this problem naturally emerge if some of the stock shares are not fixed and associated with a cost, and a holding company wants to understand whether its companies can exert an indirect control over other companies of interest.\footnote{\url{https://www.mat.unical.it/aspcomp2011/FinalProblemDescriptions/CompanyControlsOptimize}}

The possibility to mix guesses, inductive definitions, and aggregation functions opens the gates for several alternative interpretations of the expected outcome, and    thus a general semantics of aggregates and its efficient implementation have been long-standing challenges.
Already in the propositional case, checking the satisfiability of an aggregate expression is an intractable problem in general, with the \NP-complete \emph{subset sum} and \emph{subset product}
problems \cite{garjoh79} as special cases.
The practical implication of such an intractability is the use of weakened propagation procedures, 
meaning that ASP solvers \begin{rev}give up on\end{rev} the propagation of some deterministic inferences to achieve algorithmic efficiency, usually obtained by translational approaches \cite{alfage15a,alvleo15a,fapfledeie08a,ferlif02a,siniso02a}.
For example, an aggregate expression involving a sum and the equality comparison is split into a conjunction of two aggregate expressions that are true when the sum is no more and no less than the original bound.
In addition, satisfiability of aggregate expressions is in general a \emph{non-convex} function:
the expression \lstinline|#sum{1 : a; -1 : b} = 0| is true for $\emptyset$, false for \lstinline|{a}| and \lstinline|{b}|, and true again for \lstinline|{a,b}|, a pattern not shown by conjunctions of (possibly negated) atoms.
In fact, satisfiability of conjunctions of positive literals is a \emph{monotone} function (i.e., at most one transition from false to true, and no other transitions), satisfiability of conjunctions of negative literals is an \emph{anti-monotone} function (i.e., at most one transition from true to false, and no other transitions), and satisfiability of conjunctions of literals in general is a \emph{convex} function (i.e., at most one transition from false to true followed by at most one transition from true to false).
The general picture for aggregate expressions is shown in Figure~\ref{fig:convex} and further detailed in Section~\ref{subsec:monotonicity}.

\begin{figure}
    \figrule
    \centering
    
    \tikzstyle{none} = [text centered]
    \tikzstyle{set} = [draw, dashed, text centered]

    \begin{tikzpicture}
        \node at (0,0) [set](m) {
            $
                \begin{array}{rr}
                   \multicolumn{2}{r}{\textbf{monotone}} \\
                   \asum_+> & \asum_+\geq \\
                   \asum_-< & \asum_-\leq \\
                   \atimes_+> & \atimes_+\geq \\
                   \amin< & \amin\leq \\
                   \amax> & \amax\geq
                \end{array}
            $
        };
        \node at (4,0) [set](a) {
            $
                \begin{array}{rr}
                   \multicolumn{2}{r}{\textbf{anti-monotone}} \\
                   \asum_+< & \asum_+\leq \\
                   \asum_-> & \asum_-\geq \\
                   \atimes_+< & \atimes_+\leq \\
                   \amin> & \amin\geq \\
                   \amax< & \amax\leq
                \end{array}
            $
        };
        \node at (7.25,0) [none](c) {
            $
                \begin{array}{r}
                   \textbf{convex}\\
                   \asum_+= \\
                   \asum_-= \\
                   \atimes_+= \\
                   \amin= \\
                   \amax=
                \end{array}
            $
        };
        \node at (-1,-2.75) [none](n1) {
            $
                \begin{array}{r}
                   \asum_+\neq \\
                   \asum_-\neq \\
                   \atimes_+\neq \\
                   \amin\neq \\
                   \amax\neq
                \end{array}
            $
        };
        \node at (5,-2.75) [none](n2) {
            $
                \begin{array}{rrrrrr}
                    \multicolumn{6}{r}{\textbf{non-convex}} \\
                    \asum< & \asum\leq & \asum\geq & \asum> & \asum= & \asum\neq \\
                    \atimes< & \atimes\leq & \atimes\geq & \atimes> & \atimes= & \atimes\neq \\
                    \atimes_-< & \atimes_-\leq & \atimes_-\geq & \atimes_-> & \atimes_-= & \atimes_-\neq \\
                    \aavg< & \aavg\leq & \aavg\geq & \aavg> & \aavg= & \aavg\neq
                \end{array}
            $
        };
        \node [set, fit={(m) (a) (c)}, inner sep=3pt] {};
        \node [set, fit={(n1) (n2)}] {};
    \end{tikzpicture}
    \capvspace
    
    \caption{
        Aggregate expressions classified according to the monotonicity 
        of satisfiability.
        Subscript signs denote restrictions on the sign of weights;
        for example, $\asum_+$ is \asum with positive weights.
    }\label{fig:convex}
    \figrule
\end{figure}

Among the variety of semantics proposed for interpreting ASP programs with aggregates, two of them \cite{fapfle11a,ferraris11a} are implemented in widely-used ASP solvers \cite{fapfledeie08a,gekasc09c}.
The two semantics agree for programs without negation, and are thus referred indistinctly as \fflp-answer sets  in this survey.
Under this syntactic restriction, \fflp-answer sets can be defined as $\subseteq$-minimal models of a program reduct obtained by deleting rules with false bodies, as in the aggregate-free case.
Other prominent semantics \begin{rev}\cite{depebr01a,gelzha19a,liposotr10a,marrem04a,pedebr07a}\end{rev} adopt constructive procedures for checking the provability of true atoms.
\begin{rev}
While some of these semantics were originally defined via a notion of reduct in the non-disjunctive case, they can be viewed as semantics based on different extensions of the immediate consequence operator \cite{emdkow76a,lloyd87}.
Although such extensions result in\end{rev} different answer sets in general, there are cases in which answer sets according to one definition are also answer sets according to other definitions;
details on such relationships are given in Section~\ref{subsec:relationships}.

Historically, aggregates appeared in database query languages and have been available almost since the inception of database systems.
However, these aggregate constructs usually lacked generality and formal definitions.
Notably, aggregates were not present in Relational Algebra and Relational Calculus of Codd's seminal papers \cite{codd70a,codd72a}.
It was only with Klug's work \cite{klug82a} in the 1980ies that aggregate expressions were cleanly integrated into Relational Algebra and Calculus.
This integration was further refined by \citeN{ozozma87a} a few years later, which also included the ability to specify set expressions.
While none of these languages 
supported recursive definitions yet,
the Relational Calculus in these papers did, however, feature a clean formalism for defining ranges and therefore avoid unsafe queries, which can be seen as a direct precursor to safety notions in the \aspcore \begin{rev}language specification\end{rev} \cite{cafageiakakrlemarisc19a}.

Database query languages allowing for recursive definitions and aggregates were based on Datalog, essentially a deterministic fragment of ASP, in which every program is associated with a unique intended model, as for example the perfect model of Datalog programs with stratified negation \cite{apblwa87a}. Preserving the unique model property is non-trivial in presence of aggregate expressions because of nonmonotonicity of their satisfiability functions, which may lead to programs having no model, and also to programs having several equally-justifiable models.
Often in the literature, uniqueness of the intended model was guaranteed by imposing syntactic restrictions on the use of aggregates, usually resulting in some stratification of aggregations or monotonicity conditions\begin{rev}; see for example the papers by \citeN{mupira90a} and \citeN{ross94a}\end{rev}.
Other works in the literature extended the well-founded semantics to logic programs with aggregates \cite{kemstu91a,pedebr07a,gelder92a}, hence considering a third truth value to reason on undefined expressions whose 
truth or falsity is taken as unknown.
Semantics for aggregate expressions were also given in terms of translation into aggregate-free (sub)programs, and then applying the well-founded or stable model semantics to the resulting programs.
Such translations first addressed extrema predicates only, that is, \lstinline|#min| and \lstinline|#max| aggregate expressions \cite{gagrza95a,sudram91a}, and were later also presented for monotonic versions of \lstinline|#count| and \lstinline|#sum| aggregate expressions \cite{maseza13a,zayadashcoin17a}.

The first attempt to define a variant of the stable model semantics that includes aggregates was given by \citeN{kemstu91a}. The proposed generalization of the Gelfond-Lifschitz reduct was to treat aggregate expressions in the same way as negated atoms. From today's point of view, this was of course bound to be problematic, as some aggregate expressions behave like positive rather than negative literals (and some aggregate expressions behave like neither of them). As a simple example, consider the program
\begin{asp}
  a :- #sum{1 : a} > 0.
\end{asp}
which should intuitively be equivalent to
\begin{asp}
  a :- a.
\end{asp}
but will have two answer sets according to  \citeN{kemstu91a}: the expected $\emptyset$ plus the unexpected \lstinline|{a}| (as in that case the reduct results in \lstinline{a.}).
Still, this seminal work should be regarded as the starting point of what we discuss in the sequel.

The structure of this survey is as follows.
Section~\ref{sec:aggregates} introduces the language of logic programs with aggregate expressions to be studied (Section~\ref{subsec:syntax}), investigates the complexity of deciding satisfiability of aggregate expressions (Section~\ref{subsec:satisfiability}), and classifies aggregate expressions according to their monotonicity (Section~\ref{subsec:monotonicity}).
Contrary to this introduction, which used examples of symbolic programs with object variables, the main part of the survey focuses on a propositional language, where object variables are assumed to be eliminated as usual by a grounding procedure.
Section~\ref{sec:answersets} discusses model-based and construction-based answer set semantics for logic programs containing aggregates (presented in Section~\ref{subsec:model} or Section~\ref{subsec:construction}, respectively), and outlines semantic correspondences obtained for programs with aggregate expressions of particular monotonicity (Section~\ref{subsec:relationships}).
The computational complexity of common reasoning tasks is studied in Section~\ref{sec:complexity},
addressing
answer set checking (Section~\ref{subsec:check}) and answer set existence (Section~\ref{subsec:exist}).
Finally, Section~\ref{sec:discussion} discusses related work beyond the presented aggregate semantics
and properties,
including first-order semantics for symbolic programs with aggregates over object variables,
rewriting methods for turning programs containing (sophisticated) aggregates into simpler and often
aggregate-free representations, 
and tools to extend logic programs by custom aggregates.


\section{Rules and Programs with Aggregates}\label{sec:aggregates}

This section addresses basic syntactic and semantic concepts of propositional logic programs
with aggregate expressions.
Section~\ref{subsec:syntax} introduces the syntax of aggregate expressions as well as
rules and logic programs including them, and further specifies the satisfaction
of these constructs by interpretations over propositional atoms.
In Section~\ref{subsec:satisfiability}, we investigate the complexity of deciding
whether an aggregate expression is satisfiable, i.e., checking the existence of some
interpretation satisfying the aggregate expression.
Finally, Section~\ref{subsec:monotonicity} turns to monotonicity properties of
aggregate expressions, which characterize their satisfaction w.r.t.\ evolving interpretations.

\subsection{Syntax and Satisfaction of Aggregates}\label{subsec:syntax}

In order to characterize the similarities and differences between
the main aggregate semantics proposed for propositional
logic programs
\begin{rev}\cite{depebr01a,fapfle11a,ferraris11a,gelzha19a,liposotr10a,marrem04a,pedebr07a}\end{rev},%
\footnote{%
\citeN{baleme11a}, \citeN{gehakalisc15a}, and \citeN{harlif19a} provide
first-order generalizations of the aggregate semantics in \cite{fapfle11a,ferraris11a},
while \citeN{gelzha19a} as well as \citeN{pedebr07a} genuinely accommodate first-order aggregates.
Specific aggregates in the form of cardinality and weight constraints
are elaborated in \cite{ferlif02a,liuyou13a,siniso02a}.}
we consider a set \at of \emph{propositional atoms}.
An \emph{aggregate expression} has the form\pagebreak[1]
\begin{equation}\label{eq:syntax:agg}
\agg\lits{w_1:p_1,\dots,w_n:p_n}\op\bound
\end{equation}
where $n\geq 0$ and
\begin{itemize}
\item
$\agg\in\{\asum, \atimes, \aavg, \amin, \amax\}$
is the name of an \emph{aggregation function},
\item
$\bound,w_1,\dots,w_n$ are integers,
where $w_i$ is a \emph{weight}, for $1 \leq i \leq n$, and
\bound is a \emph{bound},
\item
$p_1,\dots,p_n$ are propositional atoms from \at, and
\item
$\op\in\{<,\leq,\geq,>,=,\neq\}$ is a comparison operator.%
\end{itemize}
\begin{rev}
Note that the common \acount aggregation function is a special case of \asum such that
each weight is $1$.
Similarly, \aeven and \aodd are special cases of \atimes such that each weight is $-1$, the bound is $1$, and $\op$ is $=$ or $\neq$, respectively.
More liberal aggregate concepts also account for rational or
real numbers rather than integers, formulas instead of propositional atoms only,
as well as a second comparison operator and bound \cite{cafageiakakrlemarisc19a,ferraris11a}.
\end{rev}

For an aggregate expression~$A$ as in~\eqref{eq:syntax:agg},
by $\atoms{A}=\{p_1,\dots,p_n\}$ we denote the set of
propositional atoms occurring in~$A$.

A \emph{rule} has the form
\begin{equation}\label{eq:syntax:rule}
p_1 \vee \dots \vee p_m \leftarrow A_1 \wedge \dots \wedge A_n
\end{equation}
where $m\geq 0$, $n\geq 0$,
$p_1,\dots,p_m$ are propositional atoms, and
$A_1,\dots,A_n$ are aggregate expressions.
For a rule $r$ as in \eqref{eq:syntax:rule},
let $\head{r}=\{p_1,\dots,p_m\}$ and $\body{r}=\{A_1,\dots,A_n\}$
denote the \emph{head} and \emph{body} of~$r$.
We say that $r$ is \emph{non-disjunctive} if $|\head{r}|\leq 1$, in which
case the consequent of \eqref{eq:syntax:rule} is either of the form $p_1$
or~$\bot$, the latter representing an empty disjunction, i.e., a contradiction.
The rule~$r$ is also called a \emph{constraint} if $\head{r}=\emptyset$,
and a \emph{fact} if $\body{r}=\emptyset$,
where an empty conjunction in the antecedent of~\eqref{eq:syntax:rule} is denoted by $\top$.
While the bodies of rules do not explicitly allow for (negated) propositional atoms~$p$,
they can easily be represented by aggregate expressions of the form
$\asum\lits{1:p}>0$ or $\asum\lits{1:p}<1$,
so that the syntax at hand includes disjunctive rules \cite{eitgot95a,gellif91a}.
A \emph{program} \prg is a finite set of rules, and we call \prg a
\emph{non-disjunctive} program if each $r\in\prg$ is non-disjunctive.

An \emph{interpretation} $X\subseteq\at$ is represented by the set of
its true propositional atoms.
It allows us to map each aggregate expression~$A$ of the form~\eqref{eq:syntax:agg} to
a multiset $\multi{A}{X}=[w_i \mid 1\leq i\leq n,p_i\in X]$ of weights, and then to
an expression $\eval{A}{X}$ 
where
\begin{itemize}
\item for $\agg=\asum$, $\eval{A}{X} = \sum_{w\in\multi{A}{X}}w$,
\item for $\agg=\atimes$, $\eval{A}{X} = \prod_{w\in\multi{A}{X}}w$,
\item for $\agg=\aavg$,
      $\eval{A}{X} = \left\{
           \begin{array}{@{}l@{}l@{}}
           \sum_{w\in\multi{A}{X}}w/|\multi{A}{X}|
             & \text{, if $|\multi{A}{X}| > 0$,}
           \\
           \undef 
             & \text{, if $|\multi{A}{X}| = 0$,}
           \end{array}
           \right.$
\item for $\agg=\amin$, $\eval{A}{X} = \min\{w \mid w\in\multi{A}{X}\}$, and
\item for $\agg=\amax$, $\eval{A}{X} = \max\{w \mid w\in\multi{A}{X}\}$.%
\end{itemize}
\begin{rev}
Note that the average over a (non-empty) multiset of weights
can be a rational number, e.g., $(1+2)/|[1,2]|=3/2$.    
The \undef outcome of taking the average over the
empty multiset stands for undefined,
while other aggregation functions map the empty multiset to their neutral elements:
$\sum_{w\in[]}w=0$, $\prod_{w\in[]}w=1$, $\min\{w \mid w\in[]\}=\infty$,
and $\max\{w \mid w\in[]\}=-\infty$.
\end{rev}

For
$\op\in\{<,\leq,\geq,>,=,\neq\}$,
by $X\models (\eval{A}{X} \op \bound)$ we denote that $\eval{A}{X} \op \bound$ holds,
and write $X\nmodels (\eval{A}{X} \op \bound)$ otherwise.
Note that
$X\models (\infty \geq \bound)$,
$X\models (\infty >\nolinebreak \bound)$,
$X\models (\infty \neq \bound)$, 
$X\models (-\infty < \bound)$,
$X\models (-\infty \leq \bound)$, and
$X\models (-\infty \neq \bound)$
apply independently of the (integer) bound \bound, while
$X\nmodels (\infty < \bound)$,
$X\nmodels (\infty \leq \bound)$,
$X\nmodels (\infty = \bound)$, 
$X\nmodels (-\infty \geq \bound)$,
$X\nmodels (-\infty > \bound)$,
$X\nmodels (-\infty = \bound)$, and
$X\nmodels (\undef \op \bound)$, for $\op\in\{<,\leq,\geq,>,=,\neq\}$,
are the cases in which $\eval{A}{X} \op \bound$ cannot hold.

An aggregate expression~$A$ of the form~\eqref{eq:syntax:agg} is \emph{satisfied}
by an interpretation~$X$,
also written $X\models A$, if $X\models (\eval{A}{X} \op \bound)$,
and otherwise $X\nmodels A$ denotes that $A$ is \emph{unsatisfied} by~$X$.
The body of a rule~$r$ as in~\eqref{eq:syntax:rule} is satisfied by~$X$,
written $X\models\body{r}$,
if $X\models A$ for each $A\in\body{r}$,
and unsatisfied by~$X$, indicated by $X\nmodels \body{r}$, otherwise.
Likewise, the head of~$r$ is satisfied by $X$, $X\models\head{r}$,
if $\head{r}\cap X \neq \emptyset$, and otherwise $X\nmodels\head{r}$
expresses that $\head{r}$ is unsatisfied by~$X$.
The rule $r$ is satisfied by~$X$, i.e., $X\models r$,
if $X\models \body{r}$ implies $X\models \head{r}$,
while $r$ is unsatisfied by~$X$, $X\nmodels r$, otherwise.
Note that we have $X\nmodels \head{r}$ when $r$ is a constraint,
so that $X\nmodels \body{r}$ is necessary to satisfy a constraint~$r$ by~$X$.
For a fact~$r$, $X\models \body{r}$ is instantaneous, and $r$ is satisfied by~$X$
only if $X\models \head{r}$.
Finally, $X$ is a \emph{model} of a program \prg, denoted by $X\models \prg$,
if $X\models r$ for every $r\in\prg$,
and we write $X\nmodels \prg$ otherwise.

\begin{example}\label{ex:model}
Consider the program $\prg_1$ consisting of three rules as follows:\pagebreak[1]
\begin{alignat}{1}
  \label{eq1:r1}\tag{$r_1$}
  & p \leftarrow \asum\lits{1:p,-1:q} \geq 0
  \\
  \label{eq1:r2}\tag{$r_2$}
  & p \leftarrow \asum\lits{1:q} > 0
  \\
  \label{eq1:r3}\tag{$r_3$}
  & q \leftarrow \asum\lits{1:p} > 0
\end{alignat}
Intuitively, rule $r_1$ requires $p$ to be true if $p$ is true or $q$ is false, rule $r_2$ requires $p$ true if $q$ is true, and rule $r_3$ requires $q$ true if $p$ is true.
The four interpretations over the set $\at=\{p,q\}$ of propositional atoms are
$X_1=\emptyset$, $X_2=\{p\}$, $X_3=\{q\}$, and $X_4=\{p,q\}$.
We have that $\eval{\asum\lits{1:p,-1:q} \geq 0}{X_1}=0$,
so that $X_1\models \body{$\ref{eq1:r1}$}$ but $X_1\nmodels \head{$\ref{eq1:r1}$}$.
For $X_2$ and $X_3$,
$\eval{\asum\lits{1:p} > 0}{X_2}=1$ and
$\eval{\asum\lits{1:q} > 0}{\linebreak[1]X_3}=1$ yield
$X_2\models \body{$\ref{eq1:r3}$}$  and
$X_3\models \body{$\ref{eq1:r2}$}$, while
$X_2\nmodels \head{r_3}$ and
$X_3\nmodels \head{r_2}$.
That is, neither $X_1$, $X_2$, nor $X_3$ is a model of $\prg_1$.
The remaining interpretation $X_4$ is such that
$X_4\models \body{r}$ and $X_4\models \head{r}$ for every $r\in\prg_1$,
so that $X_4$ is a model of~$\prg_1$.
\end{example}

\subsection{Satisfiability of Aggregates}\label{subsec:satisfiability}

With the syntax and satisfaction of aggregate expressions at hand,
let us turn to the complexity of deciding whether an aggregate expression~$A$ is \emph{satisfiable},
i.e., checking whether there is some interpretation $X\subseteq\at$ such that
$X\models A$.
To this end,
we first note that determining the smallest and greatest feasible outcome,
denoted by $\lb{A}$ and $\ub{A}$,
of the aggregation function in~$A$ of the form~\eqref{eq:syntax:agg} can be accomplished as follows (where $\undef$ denotes an \emph{undefined outcome}):
\begin{itemize}
\item for $\agg=\asum$,
      $\begin{array}[t]{@{}l@{}c@{}l@{}}
       \lb{A} & {} = {} & \sum_{p\in\atoms{A},\eval{A}{\{p\}}<0}\eval{A}{\{p\}}\text{,}
       \\
       \ub{A} & {} = {} & \sum_{p\in\atoms{A},\eval{A}{\{p\}}>0}\eval{A}{\{p\}}\text{,}
       \end{array}$       
\item for $\agg=\atimes$,
      let $\begin{array}[t]{@{}l@{}c@{}l@{}}
           \multicolumn{3}{@{}l@{}}{
           \begin{array}[t]{@{}r@{}c@{}l@{}}
           \pi_\pm & {} = {} & \prod_{p\in\atoms{A},\eval{A}{\{p\}}\neq 0}\eval{A}{\{p\}}\text{,}
           \\
           \pi_0   & {} = {} & \prod_{p\in\atoms{A},\eval{A}{\{p\}}= 0}\eval{A}{\{p\}}\text{, and}
           \\
           w       & {} = {} & \max\{\eval{A}{\{p\}} \mid p\in\atoms{A}, \eval{A}{\{p\}}< 0\} \text{ in}
           \end{array}}
           \\
       \lb{A} & {} = {} & 
         \left\{
         \begin{array}{@{}l@{}l@{}}
         \pi_\pm
         & \text{, if $\pi_\pm < 0$,} 
         \\
         \pi_\pm/w
         & \text{, if $\pi_\pm > 0$ and $w\neq-\infty$,} 
         \\
         \pi_0
         & \text{, if $w=-\infty$,} 
         \end{array}
         \right.
       \\
       \ub{A} & {} = {} &
         \left\{
         \begin{array}{@{}l@{}l@{}}
         \pi_\pm
         & \text{, if $\pi_\pm > 0$,} 
         \\
         \pi_\pm/w
         & \text{, if $\pi_\pm < 0$,} 
         \end{array}
         \right.
       \end{array}$
\item for $\agg=\aavg$,
      $\begin{array}[t]{@{}l@{}c@{}l@{}}
       \lb{A} & {} = {} & 
         \left\{
         \begin{array}{@{}l@{}l@{}}
         \min\{\eval{A}{\{p\}} \mid p\in\atoms{A}\}
         & \text{, if $\atoms{A}\neq\emptyset$,}
         \\
         \undef
         & \text{, if $\atoms{A}=\emptyset$,}
         \end{array}
         \right.
       \\
       \ub{A} & {} = {} &
         \left\{
         \begin{array}{@{}l@{}l@{}}
         \max\{\eval{A}{\{p\}} \mid p\in\atoms{A}\}
         & \text{, if $\atoms{A}\neq\emptyset$,}
         \\
         \undef
         & \text{, if $\atoms{A}=\emptyset$,}
         \end{array}
         \right.
       \end{array}$
\item for $\agg=\amin$,
      $\begin{array}[t]{@{}l@{}c@{}l@{}}
       \lb{A} & {} = {} & \min\{\eval{A}{\{p\}} \mid p\in\atoms{A}\}\text{,}
       \\
       \ub{A} & {} = {} & \infty\text{,}
       \end{array}$
\item for $\agg=\amax$,
      $\begin{array}[t]{@{}l@{}c@{}l@{}}
       \lb{A} & {} = {} & -\infty\text{,}
       \\
       \ub{A} & {} = {} & \max\{\eval{A}{\{p\}} \mid p\in\atoms{A}\}\text{.}
       \end{array}$
\end{itemize}
For the \asum aggregation function, the lower and upper bound are obtained
by summing over all atoms associated with a negative or positive weight, respectively.
In case of \aavg, we just pick an atom with the smallest or greatest 
weight, provided that some atom occurs in~$A$.
For \amin and \amax, either the upper or the lower bound is trivial, and
the respective counterpart is obtained by applying the aggregation function
to all weights.
The \atimes aggregation function is the most sophisticated,
as negative weights may swap the sign of its outcome.
If 
the product $\pi_\pm$ over all atoms with a non-zero weight is negative,
$\pi_\pm$ yields the lower bound,
while it 
is otherwise divided by the greatest negative weight~$w$
(i.e., smallest absolute value), provided that it exists, to achieve a negative outcome.
In the remaining case that no negative outcome is feasible, the lower bound
\begin{rev}$\pi_0$\end{rev} is zero when there is some atom with the weight zero, or it defaults to one.
Finally, the upper bound is given by $\pi_\pm$ if it is positive,
and otherwise we divide $\pi_\pm$ by the greatest negative weight~$w$ to obtain the
greatest positive outcome.

Given $\lb{A}$ and $\ub{A}$ for an aggregate expression~$A$
of the form~\eqref{eq:syntax:agg},
$\lb{A}=\ub{A}=\undef$ indicates that $A$ is unsatisfiable
regardless of the comparison operator \op,
which applies only when $A$ matches $\aavg\lits{}\op\bound$.
Otherwise,
if the comparison operator \op is $<$ or $\leq$,
we have that $A$ is satisfiable \ifaoif $\lb{A}<\bound$ or
$\lb{A}\leq\bound$ holds, respectively.
The comparison operators $>$ and $\geq$ are dual,
and an aggregate expression~$A$ including one of them is satisfiable \ifaoif
either $\ub{A}>\bound$ or
$\ub{A}\geq\bound$ holds.
In case \op is $\neq$, we have that $A$ is satisfiable \ifaoif
$\lb{A}\neq\bound$ or $\ub{A}\neq\bound$ holds.
These considerations show that checking the satisfiability of an
aggregate expression with a comparison operator other than $=$ is tractable, i.e.,
in the complexity class~\poly.
The situation gets more involved when \op is $=$, where the aggregation functions
\amin and \amax still allow for tractable decisions:
an aggregate expression~$A$ with \agg either \amin or \amax and $=$ for \op
is satisfiable \ifaoif $\eval{A}{\{p\}}=\bound$ for some $p\in\atoms{A}$.
Unlike that, for $\agg=\asum$ and $\agg=\atimes$, deciding whether there
is some (non-empty) subset~$X$ of $\atoms{A}$ such that $\eval{A}{X}=\bound$
amounts to the \NP-complete \emph{subset sum} or \emph{subset product}
problem \cite{garjoh79}.
Concerning $\agg=\aavg$,
subset sum can be reduced to checking whether an aggregate expression~$A$ with bound $0$ is satisfiable:
$\asum\lits{w_1:p_1,\dots,w_n:p_n}=\bound$
with positive integers $w_0,\dots,w_n$ is satisfiable \ifaoif
$\aavg\lits{w_1:p_1,\dots,w_n:p_n,-\bound:p}=0$ is satisfiable,
where $p\in\at$ is some new atom.
As constants like $\bound+1$ can be added to each weight and the bound
of an aggregate expression with \aavg,
we have \NP-completeness of checking whether $A$ with $\agg\in\{\asum,\atimes,\aavg\}$
and $=$ for \op is satisfiable,
which applies regardless of whether the included weights and bound are restricted to
be positive or negative integers only.

\begin{example}\label{ex:satisfiable}
Consider aggregate expressions constructed as follows:\pagebreak[1]
\begin{alignat}{1}
  \label{eq2:A1}\tag{$A_1$}
  \asum\lits{1:p_1,3:p_2,3:p_3,-4:p_4} \op \bound
  \\
  \label{eq2:A2}\tag{$A_2$}
  \atimes\lits{0:p_1,3:p_2,-2:p_3,-4:p_4} \op \bound
  \\
  \label{eq2:A3}\tag{$A_3$}
  \aavg\lits{1:p_1,2:p_2,3:p_3,6:p_4} \op \bound
  \\
  \label{eq2:A4}\tag{$A_4$}
  \amin\lits{0:p_1,3:p_2,-2:p_3,-4:p_4} \op \bound
  \\
  \label{eq2:A5}\tag{$A_5$}
  \amax\lits{1:p_1,3:p_2,3:p_3,-4:p_4} \op \bound
\end{alignat}
By summing up negative or positive weights, respectively,
we get $\lb{A_1}=-4$ and $\ub{A_1}=7$.
Hence, $A_1$ is satisfiable
for ${\op}={<}$ and any bound \bound greater than $-4$,
for ${\op}={\leq}$ and \bound not smaller than $-4$,
for ${\op}={\geq}$ and \bound not greater than $7$,
for ${\op}={>}$ and \bound smaller than $7$, and
for ${\op}={\neq}$ with any bound~\bound.
When \op is $=$, satisfiability is more intricate,
and one can check that $A_1$ is satisfiable for bounds \bound other than
$-2$ and $5$ in the range from $\lb{A_1}=-4$ to $\ub{A_1}=7$.

The product of non-zero weights, two of which are negative,
for $A_2$ is $24$.
To obtain the lower bound for $A_2$, we have thus to divide by
the greatest negative weight $-2$, leading to $\lb{A_2}=-12$,
while $\ub{A_2}=24$.
For $\op\in\{<,\leq,\geq,>,\neq\}$,
satisfiability is determined similar to $A_1$,
yet w.r.t.\ the lower and upper bound for $A_2$.
When \op is $=$, we have that $A_2$ is satisfiable for
$\bound\in\{-12,\linebreak[1]-6,\linebreak[1]-4,\linebreak[1]-2,
 \linebreak[1]0,\linebreak[1]1,\linebreak[1]3,\linebreak[1]8,\linebreak[1]24\}$,
and unsatisfiable otherwise.

Concerning $A_3$, we get $\lb{A_3}=1$ and $\ub{A_3}=6$,
and satisfiability for $\op\in\{<,\leq,\geq,>,\neq\}$
w.r.t.\ these bounds is analogous.
As $1$, $2$, $3$, $4$, and $6$ are the feasible integer outcomes,
$A_3$ with $=$ for \op is satisfiable for such outcomes as \bound,
and unsatisfiable for any other bound~\bound.

Turning to $A_4$ and $A_5$,
the lower and upper bounds are
$\lb{A_4}=-4$, $\ub{A_4}=\infty$, $\lb{A_5}=-\infty$, and $\ub{A_5}=3$.
For $\op\in\{<,\leq,\geq,>,\neq\}$,
satisfiability is determined similar to $A_1$,
where $A_4$ is satisfiable for ${\op}={\geq}$, ${\op}={>}$, and ${\op}={\neq}$
regardless of the bound \bound,
and likewise $A_5$ for ${\op}={<}$, ${\op}={\leq}$, and ${\op}={\neq}$.
When \op is ${=}$, we have that $A_4$ and $A_5$ are satisfiable for
bounds \bound matching their contained weights, i.e.,
$\bound\in\{-4,-2,0,3\}$ or $\bound\in\{-4,1,3\}$, respectively.
\end{example}

\subsection{Monotonicity of Aggregates}\label{subsec:monotonicity}

A relevant property related to satisfiability concerns the 
(non)monotonicity of aggregate expressions (cf.\ \citeNP{fapfle11a,liposotr10a}) \begin{rev}and the notion of convex aggregates introduced by \citeN{liutru06a}\end{rev}.
An aggregate expression~$A$ is \emph{convex} if $X \subseteq Y$, $X \models A$ and $Y \models A$ imply $Z \models A$ for any interpretation~$Z$ between $X$ and $Y$, that is,
$X\models A$ and $Y\models A$ yield $Z\models A$ for all $X\subseteq Z\subseteq Y$.
Otherwise, the aggregate expression is \emph{non-convex}.
Two frequent special cases of convexity are given by
\emph{monotone} and \emph{anti-monotone} aggregate expressions,
for which $X\models A$ implies that $\at\models A$ or
$\emptyset\models A$, respectively.
That is, a monotone aggregate expression~$A$ remains satisfied when adding
(true) propositional atoms to an interpretation~$X$ satisfying~$A$,
while the satisfaction of an anti-monotone~$A$ is preserved when
propositional atoms from $X$ are falsified.

%
Specific monotonicity properties that follow from the aggregation
functions and comparison operators of aggregate expressions are summarized
in Figure~\ref{fig:convex} in the introduction.
For the \asum and \atimes aggregation functions,
restrictions to positive or negative weights and bounds 
affect monotonicity, and the subscripts $_+$ and $_-$ indicate
such particular conditions.
Unlike that, the signs of weights and bounds are immaterial for
\aavg, \amin, and \amax, and we also disregard special cases like the
unsatisfiability of $\aavg\lits{}\op\bound$, which makes aggregate expressions 
with \aavg over the empty multiset monotone, anti-monotone, and convex.

While aggregate expressions with \asum are in general non-convex regardless of the comparison
operator, restrictions to positive or negative weights and bounds make them convex
for comparison operators other than $\neq$,
given that the sum of weights can merely increase or decrease when 
propositional atoms are added to an interpretation.
More specifically, positive weights and bounds lead to an anti-monotone aggregate expression
for $<$ or $\leq$ as comparison operator, and to a monotone aggregate expression for $\geq$
or $>$, where dual properties apply in case of negative weights and bounds only.
For either restriction,
the comparison operator~$\neq$ does not lead to more regular monotonicity though
because a respective aggregate expression may be satisfied by the empty interpretation~$\emptyset$,
become unsatisfied when atoms whose weights sum up to the bound~\bound are made true,
and then get satisfied again by adding further propositional atoms to the interpretation.

Aggregate expressions with \atimes are generally non-convex, even if weights and bounds are
restricted to be negative only.
This is due to the condition that negative weights swap the sign of the outcome,
so that satisfaction may switch back and forth for any comparison operator.
With positive weights and bounds (excluding zero),
the outcome increases when more propositional atoms are included in an interpretation.
Hence, such aggregate expressions yield similar monotonicity as \asum with positive weights and bounds,
and entries for $\asum_+$ and $\atimes_+$ in Figure~\ref{fig:convex} match.

For the remaining aggregation functions, \aavg, \amin, and \amax,
the signs of weights do not make a difference regarding monotonicity.
In fact, the outcome of \aavg may increase or decrease when propositional atoms
are added to an interpretation, so that corresponding aggregate expressions are generally
non-convex regardless of the comparison operator.
The outcomes of \amin and \amax monotonically decrease or increase, respectively,
with growing interpretations, leading to (an\-ti\nobreakdash-)\-mo\-no\-to\-ni\-ci\-ty for the comparison
operators $<$ or $\leq$ and $\geq$ or $>$, as well as convexity for~$=$.
Similar to the other aggregation functions, 
no matter whether weights and bounds are positive, negative, or unrestricted,
$\neq$ as comparison operator makes aggregate expressions with \amin and \amax
non-convex in general,
as they are satisfied by the empty interpretation~$\emptyset$, may become
unsatisfied and then satisfied again when the interpretation is extended
by propositional atoms.

\begin{example}\label{ex:monotonicity}
The aggregate expression\pagebreak[1]
\begin{alignat}{1}
  \label{eq3:A6}\tag{$A_6$}
  \asum\lits{1:p_1,2:p_2,2:p_3,3:p_4} \op 5
\end{alignat}
is monotone for ${\op}={\geq}$ or ${\op}={>}$,
anti-monotone for ${\op}={<}$ or ${\op}={\leq}$,
convex when \op is $=$, and
non-convex with $\neq$ for \op.
The three mutually incomparable interpretations $\{p_1,p_2,p_3\}$,
$\{p_2,p_4\}$, and $\{p_3,p_4\}$ satisfying $A_6$ with $=$ for \op
meet the condition for convexity, but neither the additional requirements
for monotonicity nor anti-monotonicity.
Moreover, they witness the non-convexity of $A_6$ when \op is $\neq$,
as there are interpretations with both fewer and more propositional atoms such that
their (positive) weights do not sum up to the bound~$5$ and satisfy~$A_6$
for ${\op}={\neq}$.
\end{example}


\section{Answer Set Semantics of Aggregates}\label{sec:answersets}

This section characterizes the main answer set semantics proposed for
logic programs with aggregate expressions
\begin{rev}\cite{depebr01a,fapfle11a,ferraris11a,gelzha19a,liposotr10a,marrem04a,pedebr07a}\end{rev}.
Their common essence is to extend the notion of provability of
the true propositional atoms in a model to programs that include aggregate expressions.
In Section~\ref{subsec:model}, we review aggregate semantics based on
$\subseteq$-minimal models of a program reduct.
Section~\ref{subsec:construction} then gathers several constructive approaches
to capture the provability of true atoms.
Correspondences between different aggregate semantics w.r.t.\ the monotonicity of
aggregate expressions are analyzed in Section~\ref{subsec:relationships}.

\subsection{Model-based Semantics}\label{subsec:model}

In line with \cite{gellif88b,gellif91a},
we let the \emph{reduct} of a program \prg relative to
an interpretation $X\subseteq\at$ be
$\reduct{\prg}{X}=\{r\in\prg \mid X\models\body{r}\}$.
In other words, the reduct is obtained from \prg by dropping rules
whose bodies are unsatisfied by~$X$.
(Note that the original definition of reduct provided by \citeN{gellif88b} also removes negative literals, which are not part of the language considered in this paper.)
Hence, we have that $X$ is a model of \prg \ifaoif $X$ is a model of $\reduct{\prg}{X}$.

Different answer set semantics inspect the reduct in specific ways to
check the provability of true atoms.
We note that our reduct notion matches the definitions given by \citeN{fapfle11a} and by \citeN{ferraris11a}, as it does not eliminate
false atoms or falsified expressions, respectively.
The latter would happen for negated atoms with the seminal Gelfond-Lifschitz
reduct \cite{gellif88b,gellif91a}, and entirely with the reduct by \citeN{ferraris11a}.
However, the syntax of aggregate expressions and rules in
\eqref{eq:syntax:agg}--\eqref{eq:syntax:rule} is chosen such that specific
reducts do not make a difference regarding answer sets.

The model-theoretic approaches by 
\begin{rev}Faber, Pfeifer and Leone \citeyear{fapfle11a}\end{rev}
and \citeN{ferraris11a} agree in
focusing on $\subseteq$-minimal models of the reduct $\reduct{\prg}{X}$
of a program \prg relative to an interpretation $X\subseteq\at$.
We use the convention to indicate particular semantics by their proposers,
and call $X$ an \emph{\fflp-answer set} of \prg if $X$ is a $\subseteq$-minimal
model of $\reduct{\prg}{X}$.
That is, an \fflp-answer set of \prg fulfills two conditions:
\begin{itemize}
\item $X$ is a model of \prg, i.e., $X\models \prg$, and
\item for any interpretation $Y\subset X$, we have that $Y\nmodels\reduct{\prg}{X}$.
\end{itemize}
While the first condition is equivalent to $X\models \reduct{\prg}{X}$,
exchanging $Y\nmodels\reduct{\prg}{X}$ for $Y\nmodels \prg$ in the second
condition would yield a different semantics.

\begin{example}\label{ex:unsatisfiable}
Consider the program $\prg_2$ containing the following two rules:\pagebreak[1]
\begin{alignat}{1}
  \label{eq4:r4}\tag{$r_4$}
  & p \leftarrow \asum\lits{1:p} > 0
  \\
  \label{eq4:r5}\tag{$r_5$}
  & p \leftarrow \asum\lits{1:p} < 1
\end{alignat}
Intuitively, rule $r_4$ requires $p$ true if $p$ is true, and rule $r_5$ requires $p$ true if $p$ is false.
The interpretation $X_1=\emptyset$ is not a model of $\prg_2$
because $\eval{\asum\lits{1:p} < 1}{X_1}=0$ yields that
$X_1\models \body{$\ref{eq4:r5}$}$ but $X_1\nmodels \head{$\ref{eq4:r5}$}$.
For $X_2=\{p\}$, we have that $\eval{\asum\lits{1:p} < 1}{X_2}=1$,
so that $X_2\nmodels \body{$\ref{eq4:r5}$}$.
Hence, $\reduct{\prg_2}{X_2}$ consists of the rule \ref{eq4:r4} only.
Since $X_1\subset X_2$ is such that $\eval{\asum\lits{1:p} > 0}{X_1}=0$,
$X_1\nmodels \body{$\ref{eq4:r4}$}$, and $X_1\models \reduct{\prg_2}{X_2}$,
the interpretation $X_1$ shows that $X_2$ is not a $\subseteq$-minimal
model of $\reduct{\prg_2}{X_2}$.
As a consequence, the program $\prg_2$ does not have any \fflp-answer set.
However, note that $X_2$ is the $\subseteq$-minimal model of $\prg_2$
in view of $X_2\models \prg_2$ and $X_1\nmodels \prg_2$.
\end{example}

\begin{example}\label{ex:minmodel}
Reconsider the program~$\prg_1$ from Example~\ref{ex:model}:\pagebreak[1]
\begin{alignat}{1}
  \tag{\ref{eq1:r1}}
  & p \leftarrow \asum\lits{1:p,-1:q} \geq 0
  \\
  \tag{\ref{eq1:r2}}
  & p \leftarrow \asum\lits{1:q} > 0
  \\
  \tag{\ref{eq1:r3}}
  & q \leftarrow \asum\lits{1:p} > 0
\end{alignat}
As checked in Example~\ref{ex:model},
we have that $X_4=\{p,q\}$ is the unique model of $\prg_1$
(over the set $\at=\{p,q\}$ of propositional atoms).
Given that $X_4\models \body{r}$ for every $r\in\prg_1$,
we obtain $\reduct{\prg_1}{X_4}=\prg_1$,
which shows that $X_4$ is an \fflp-answer set of $\prg_1$.
\end{example}

\begin{example}\label{ex:subsum}
Consider the program $\prg_3$ consisting of the following rules:\pagebreak[1]
\begin{alignat}{1}
  \label{eq6:r6}\tag{$r_6$}
  & x_1 \leftarrow \asum\lits{1:y_1} < 1
  \\
  \label{eq6:r7}\tag{$r_7$}
  & y_1 \leftarrow \asum\lits{1:x_1} < 1
  \\
  \label{eq6:r8}\tag{$r_8$}
  & x_2 \leftarrow \asum\lits{1:y_2} < 1
  \\
  \label{eq6:r9}\tag{$r_9$}
  & y_2 \leftarrow \asum\lits{1:x_2} < 1
  \\
  \label{eq6:r10}\tag{$r_{10}$}
  & z_1 \leftarrow \asum\lits{1:p} > 0
  \\
  \label{eq6:r11}\tag{$r_{11}$}
  & z_2 \leftarrow \asum\lits{1:p} > 0
  \\
  \label{eq6:r12}\tag{$r_{12}$}
  & p \leftarrow \asum\lits{1:y_1,2:y_2,2:z_1,3:z_2} \neq 5
  \\
  \label{eq6:r13}\tag{$r_{13}$}
  & \bot \leftarrow \asum\lits{1:p} < 1
\end{alignat}
As will be clarified later in Section~\ref{sec:complexity} and Example~\ref{ex:exist}, $\prg_3$ \begin{rev}(under \fflp-, \lpst-, or \pdb-answer set semantics)\end{rev} encodes an instance of \emph{generalized subset sum} \cite{bekalaplry02a}, specifically
\begin{alignat*}{1}
\exists y_1 y_2 \forall z_1 z_2 (1 \cdot y_1 + 2 \cdot y_2 + 2 \cdot z_1 + 3 \cdot z_2 \neq 5)\text{.}
\end{alignat*}
Intuitively, rules $r_6$ and $r_7$ are intended to select exactly one of $x_1$ and $y_1$, and similarly rules $r_8$ and $r_9$ are intended to select exactly one of $x_2$ and $y_2$;
these rules represent the existential variables of the generalized subset sum instance.
Rules $r_{10}$ and $r_{11}$ require $z_1$ and $z_2$ to be true if $p$ is true;
these rules represent the universal variables of the generalized subset sum instance.
Rule $r_{13}$ enforces $p$ true, but does not provide a justification for it, which instead must be provided by rule $r_{12}$, encoding the inequality of the generalized subset sum instance.
\begin{rev}
Note that $\prg_3$ follows the \emph{saturation technique} by \citeN{eitgot95a} to express that $z_1$ and $z_2$ are universally quantified:
$z_1$ and $z_2$ must be true because of~$p$ 
(they are saturated),
while $p$ (and $z_1$, $z_2$) are provable \ifaoif
the universal quantification holds w.r.t.\ the chosen (true) subset of $\{y_1,y_2\}$.
\end{rev}

Let us focus on interpretations that are potential candidates for
\fflp-answer sets of $\prg_3$.
In view of the constraint \ref{eq6:r13},
each model of $\prg_3$ must include the atom~$p$,
and then the atoms $z_1$ and $z_2$ because of the rules
\ref{eq6:r10} and \ref{eq6:r11}.
Regarding $x_i$ and $y_i$ for $1\leq i\leq 2$,
having both atoms false yields unsatisfied rules \ref{eq6:r6} and \ref{eq6:r7} or
\ref{eq6:r8} and \ref{eq6:r9},
while both atoms true lead to a reduct excluding \ref{eq6:r6} and \ref{eq6:r7} or
\ref{eq6:r8} and \ref{eq6:r9}.
In the latter case, falsifying $x_i$ and $y_i$ gives a smaller model of the reduct,
so that the original interpretation cannot be an \fflp-answer set of $\prg_3$.

The previous considerations leave the models
$X_1=\{x_1,x_2,z_1,z_2,p\}$,
$X_2=\{x_1,y_2,z_1,z_2,p\}$,
$X_3=\{y_1,x_2,z_1,z_2,p\}$, and
$X_4=\{y_1,y_2,z_1,z_2,p\}$ as potential
\fflp-answer sets of $\prg_3$.
For each of these interpretations,
the reduct includes either of the rules \ref{eq6:r6} or \ref{eq6:r7}
as well as \ref{eq6:r8} or \ref{eq6:r9},
so that falsifying an atom $x_i$ or $y_i$ leads to unsatisfaction of the
rule with the atom in the head.
A smaller model of the reduct must thus falsify $p$ along with an
appropriate combination of the atoms $z_1$ and $z_2$,
in order to establish the unsatisfaction of $\body{$\ref{eq6:r12}$}$.
Note that the reduct does not include the constraint \ref{eq6:r13},
which enables the falsification of $p$, provided that $\body{$\ref{eq6:r12}$}$ is unsatisfied.

For $X_1$,
the observation that 
$\eval{\asum\lits{1:y_1,\linebreak[1]2:y_2,\linebreak[1]2:z_1,\linebreak[1]3:z_2} \neq 5}{X_1}=5$ yields 
$X_1\nmodels \body{$\ref{eq6:r12}$}$ and $\text{\ref{eq6:r12}}\notin\reduct{\prg_3}{X_1}$,
so that $Y_1\models \reduct{\prg_3}{X_1}$,
for each $\{x_1,x_2\}\subseteq Y_1\subseteq X_1\setminus\{p\}$,
disproves $X_1$ to be an \fflp-answer set of $\prg_3$.
Regarding $X_2$ and $X_4$, we have that
$Y_2=X_2\setminus\{z_1,p\}=\{x_1,y_2,z_2\}$ and
$Y_4=X_4\setminus\{z_2,p\}=\{y_1,y_2,z_1\}$ furnish 
counterexamples satisfying the respective reduct
in view of $Y_2\nmodels \body{$\ref{eq6:r12}$}$ and $Y_4\nmodels \body{$\ref{eq6:r12}$}$.
It thus remains to investigate $X_3$, where
$\reduct{\prg_3}{X_3}=\{\text{\ref{eq6:r7}},\text{\ref{eq6:r8}},\text{\ref{eq6:r10}},\text{\ref{eq6:r11}},\text{\ref{eq6:r12}}\}$ and
$\eval{\asum\lits{1:y_1,\linebreak[1]2:y_2,\linebreak[1]2:z_1,\linebreak[1]3:z_2} \neq 5}{X_3}=6$.
That is, an interpretation $Y_3\subset X_3$ with
$\eval{\asum\lits{1:\nolinebreak y_1,\linebreak[1]2:y_2,\linebreak[1]2:z_1,\linebreak[1]3:z_2} \neq 5}{Y_3}=5$
can only be obtained by falsifying $y_1$, which leads to $Y_3\nmodels \text{\ref{eq6:r7}}$.
In turn, $Y\models \body{$\ref{eq6:r12}$}$ is the case for
every model $Y\subseteq X_3$ of $\reduct{\prg_3}{X_3}$,
leaving $Y=X_3$ as the unique such interpretation.
Hence, we conclude that $X_3$ is a $\subseteq$-minimal model of $\reduct{\prg_3}{X_3}$
and thus an \fflp-answer set of $\prg_3$.
\end{example}

\subsection{Construction-based Semantics}\label{subsec:construction}

While the aggregate semantics by \citeN{fapfle11a} and \citeN{ferraris11a} build on
$\subseteq$-minimal models of a reduct, without considering the calculation
of such models in the first place,
the semantics given by \citeN{gelzha19a}, Liu, Pontelli, Son and Truszczy{\'n}ski \citeyear{liposotr10a}, \citeN{marrem04a}, and \begin{rev}Denecker, Pelov and Bruynooghe \citeyear{depebr01a}\end{rev}
focus on constructive characterizations for checking the provability of true atoms.
%
\begin{rev}
In particular, the semantics by \citeN{gelzha19a} and \citeN{liposotr10a},
referred to as \gz- and \lpst-answer set semantics
indicated by their proposers,
are based on a monotone fixpoint construction to check the stability of an answer set candidate $X$, which amounts to testing whether all atoms of $X$ (and only those) can be constructed/derived from the rules, assuming that atoms not contained in $X$ are fixed to false.
The family of semantics by \citeN{pedebr07a},
defined in terms of approximation fixpoint theory, 
provides a strongly related approach:
for each suitable choice of a three-valued truth assignment function 
mapping each aggregate expression 
to true, false, or unknown w.r.t.\ a
partial interpretation, 
the framework induces an answer set, a well-founded, and a Kripke-Kleene semantics.
As shown by \citeN{DBLP:journals/corr/abs-2104-14789},
the particular truth assignment functions for
\emph{trivial} and \emph{ultimate} approximating aggregates
\cite{pedebr07a}
lead exactly to the \gz- or \lpst-answer set semantics,
respectively.
In addition, \citeN{pedebr07a} proposed a truth assignment function for
\emph{bound} approximating aggregates, based on evaluating lower and upper bounds like $\lb{A}$ and $\ub{A}$ in Section~\ref{subsec:satisfiability}, which
achieves provability for more answer set candidates than the
\gz-answer set semantics without increasing the computational complexity.
%
Given such close relationships, we do not separately address 
trivial, ultimate, and bound approximating aggregates,
but focus on the so-called \emph{ultimate semantics},
originally introduced in earlier work \cite{depebr01a}, as 
a complementary instance of the framework by \citeN{pedebr07a}.%
\end{rev}

\begin{rev}In the following, we describe the constructions characterizing the aforementioned semantics\end{rev} by dedicated extensions of the
well-known immediate consequence operator $\tp{\prg}$ \cite{emdkow76a,lloyd87}
to programs with aggregate expressions.
We start by specifying particular notions of satisfaction,
again indicated by the proposers of a semantics, of an aggregate expression $A$ relative to two
interpretations $Y\subseteq \at,X\subseteq \at$, where $X$ is an answer set candidate and $Y$ is the reconstructed model:
\begin{itemize}
\item $(Y,X) \sat{\gz} A$ if $X\models A$ and $\atoms{A}\cap Y = \atoms{A}\cap X$,
\item $(Y,X) \sat{\lpst} A$ if $Z\models A$ for every interpretation $Y\subseteq Z\subseteq X$, and
\item $(Y,X) \sat{\mr} A$ if $X\models A$ and there is some interpretation $Z\subseteq Y$ such that $Z\models A$. 
\end{itemize}
When $Y\subseteq X$, the satisfaction relations $\sat{\gz}$, $\sat{\lpst}$, and $\sat{\mr}$
can apply only if $X\models A$, which parallels the idea of a reduct relative to~$X$.
Their second purpose is to determine partial interpretations $(Y,X)$,
where the truth of atoms in $X\setminus Y$ is considered unknown,
for which an aggregate expression $A$ should be regarded as provably true:
$(Y,X) \sat{\gz} A$ expresses that no unknown atoms from $X\setminus Y$ occur in~$A$,
$(Y,X) \sat{\lpst} A$ means that $A$ is satisfied no matter which unknown atoms
are taken to be true or false, respectively, and
$(Y,X) \sat{\mr} A$ merely requires the satisfaction of~$A$ by some subset of the
true atoms in~$Y$.
Unlike that, the 
ultimate semantics
relies on the conventional satisfaction of aggregate expressions, as introduced in
Section~\ref{subsec:satisfiability}. %
Similar to $X\models \body{r}$, we extend $\sat{\any}$, for $\any\in\{\gz,\lpst,\mr\}$,
to the body of a rule~$r$ by writing $(Y,X) \sat{\any} \body{r}$
if $(Y,X) \sat{\any} A$ for each $A\in\body{r}$,
and indicate that $(Y,X) \unsat{\any} A$, for some $A\in\body{r}$,
by $(Y,X) \unsat{\any} \body{r}$.

\begin{example}\label{ex:interpretations}
Let us investigate the aggregate expression\pagebreak[1]
\begin{alignat}{1}
  \label{eq7:A7}\tag{$A_7$}
  \asum\lits{1:y_1,2:y_2,2:z_1,3:z_2} \neq 5
\end{alignat}
along with the interpretation $X=\{y_1,z_1,z_2\}$,
where $\eval{A_7}{X}=6$ and thus $X\models A_7$.
The condition for $(Y,X) \sat{\gz} A_7$ additionally requires
$\atoms{A_7}\cap Y=\atoms{A_7}\cap X=\{y_1,z_1,z_2\}$,
so that $Y=X=\{y_1,z_1,z_2\}$ is the only subset of $X$
for which $\sat{\gz}$ applies.
For identifying interpretations $Y\subseteq X$ such that $(Y,X) \sat{\lpst} A_7$,
observe that $\eval{A_7}{\{z_1,z_2\}}=5$ and $\{z_1,z_2\}\nmodels A_7$.
That is, any subset of $X$ that excludes $y_1$ can be filled up to the interpretation
$Z=\{z_1,z_2\}$ for which $A_7$ is unsatisfied.
In turn, such a $Z$ does not exist for interpretations $\{y_1\}\subseteq Y\subseteq X$,
and we have $(Y,X) \sat{\lpst} A_7$ for each subset $Y$ of $X$ that includes $y_1$.
Regarding $(Y,X) \sat{\mr} A_7$, it is sufficient to observe that
$\eval{A_7}{\emptyset}=0$ and $\emptyset\models A_7$,
so that $\sat{\mr}$ applies for every subset $Y$ of~$X$.
\end{example}

Provided that $Y\subseteq X\subseteq \at$, 
we have that $(Y,X) \sat{\gz} A$ implies $(Y,X) \sat{\lpst} A$,
and also that $(Y,X) \sat{\lpst} A$ implies $(Y,X) \sat{\mr} A$.
Moreover, $(Y,X) \sat{\any} A$ yields $(Z,X) \sat{\any} A$
for interpretations $Y\subseteq Z\subseteq X$ and $\any\in\{\gz,\lpst,\mr\}$.
These properties carry forward to the body $\body{r}$ of a rule~$r$.

Given the specific satisfaction relations,
we now formulate \emph{immediate consequence operators}
characterizing semantics based on constructions
for a program \prg and interpretations $Y\subseteq \at,\linebreak[1]X\subseteq\nolinebreak \at$:%
\begin{alignat*}{1}
\tpc{\any}{\prg}{X}{Y} & {} = \text{$\bigcup$}_{r\in\prg,(Y,X)\sat{\any}\body{r}}\head{r}
\text{, for $\any\in\{\gz,\lpst,\mr\}$,}
\\
\tpc{\pdb}{\prg}{X}{Y} & {} = \text{$\bigcap$}_{Y\subseteq Z\subseteq X}
                              \left(
                              \text{$\bigcup$}_{r\in\prg,Z\models\body{r}}\head{r}
                              \right)
\text{.}
\end{alignat*}
Each operator $\tpx{\any}{\prg}{X}$,
for $\any\in\{\gz,\lpst,\mr,\pdb\}$,
joins the heads of rules such that the underlying satisfaction relation
applies to their bodies.
This would yield unintended results for (proper) disjunctive rules
like $p\vee q\leftarrow \top$,
having $\{p\}$ and $\{q\}$ as its standard answer sets \cite{gellif91a},
while any of the operators $\tpx{\any}{\prg}{X}$ would produce the non-minimal model $\{p,q\}$.
In fact, such immediate consequence operators have been devised with
non-disjunctive rules in mind,
and in the following we assume that $|\head{r}|\leq 1$ for rules~$r$
under consideration.
\begin{rev}For the sake of completeness, we point out that the semantics by \citeN{gelzha19a}, \citeN{liposotr10a}, and \citeN{marrem04a} additionally
allow for aggregate expressions in heads of rules,
understood as choice constructs \cite{siniso02a},
and \citeN{gelzha19a} also handle disjunctive rules,
as their original concept builds on a specific reduct rather than
an operational characterization.\end{rev}

The operators $\tpx{\any}{\prg}{X}$ 
have in common that $\tpc{\any}{\prg}{X}{Y} \subseteq \tpc{\any}{\prg}{X}{Z}$
when $Y\subseteq Z\subseteq X$.
Considering $\tpx{\pdb}{\prg}{X}$, this is because the intersection over
interpretations between $Z$ and $X$ involves fewer elements than with~$Y$,
while $(Y,X)\sat{\any}\body{r}$ implies $(Z,X)\sat{\any}\body{r}$ for the
remaining operators $\tpx{\any}{\prg}{X}$ and any rule $r\in\prg$.
Provided that $X$ is a model of \prg, in which case
$\tpc{\gz}{\prg}{X}{X} = \tpc{\lpst}{\prg}{X}{X} =
 \tpc{\mr}{\prg}{X}{X} = \tpc{\pdb}{\prg}{X}{X} \subseteq X$,
the least fixpoint of $\tpx{\any}{\prg}{X}$ 
is thus guaranteed to exist and can be constructed as follows:
\begin{alignat*}{1}
\tpu{\any}{\prg}{X}{0}   & {} = \emptyset
\text{,}
\\
\tpu{\any}{\prg}{X}{i+1} & {} = \tpc{\any}{\prg}{X}{\tpu{\any}{\prg}{X}{i}}
\text{.}
\end{alignat*}
That is, starting from the empty interpretation~$\emptyset$ for
$\tpu{\any}{\prg}{X}{0}$,
the iterated application of $\tpx{\any}{\prg}{X}$ leads to
$\tpu{\any}{\prg}{X}{i+1} = \tpc{\any}{\prg}{X}{\tpu{\any}{\prg}{X}{i}}
                          = \tpu{\any}{\prg}{X}{i}$
for some $i\geq 0$, and we denote this \emph{least fixpoint} by $\tpu{\any}{\prg}{X}{\infty}$.
For $\any\in\{\gz,\lpst,\mr,\pdb\}$,
we call $X$ a \emph{\any-answer set} of \prg if $X$ is a model of \prg
such that $\tpu{\any}{\prg}{X}{\infty} = X$.

\begin{example}\label{ex:pdb}
Reconsider the program $\prg_2$ from Example~\ref{ex:unsatisfiable}:\pagebreak[1]
\begin{alignat}{1}
  \tag{\ref{eq4:r4}}
  & p \leftarrow \asum\lits{1:p} > 0
  \\
  \tag{\ref{eq4:r5}}
  & p \leftarrow \asum\lits{1:p} < 1
\end{alignat}
We have that $X=\{p\}$ is the unique model of $\prg_2$ (over $\at=\{p\}$).
Then, $\eval{\asum\lits{1:p} < 1}{X}=1$ yields $X\nmodels \body{$\ref{eq4:r5}$}$
and $(\emptyset,X) \unsat{\any} \body{$\ref{eq4:r5}$}$ for $\any\in\{\gz,\lpst,\mr\}$.
Moreover, $(\emptyset,X) \unsat{\any} \body{$\ref{eq4:r4}$}$ follows from
$\eval{\asum\lits{1:p} > 0}{\emptyset}=0$ and $\emptyset\nmodels \body{$\ref{eq4:r4}$}$.
As a consequence, for $\any\in\{\gz,\lpst,\mr\}$,
we obtain $\tpu{\any}{\prg_2}{X}{\infty}=\tpu{\any}{\prg_2}{X}{0}=\emptyset$,
so that $X$ is not a \any-answer set of $\prg_2$.

Unlike that, $\emptyset\models \body{$\ref{eq4:r5}$}$ and
$X\models \body{$\ref{eq4:r4}$}$ with
$\head{$\ref{eq4:r5}$} = \head{$\ref{eq4:r4}$} = \{p\}$ lead to
$\tpu{\pdb}{\prg_2}{X}{\infty} = \tpu{\pdb}{\prg_2}{X}{\nolinebreak 1} =
 \tpc{\pdb}{\prg_2}{X}{\emptyset} = \{p\} = X$.
Hence, we conclude that $X$ is a \pdb-answer set of $\prg_2$.%
\end{example}

\begin{example}\label{ex:fflp}
Recall from Example~\ref{ex:minmodel}
that $X=\{p,q\}$ is the unique model and
\fflp-answer set of~$\prg_1$:\pagebreak[1]
\begin{alignat}{1}
  \tag{\ref{eq1:r1}}
  & p \leftarrow \asum\lits{1:p,-1:q} \geq 0
  \\
  \tag{\ref{eq1:r2}}
  & p \leftarrow \asum\lits{1:q} > 0
  \\
  \tag{\ref{eq1:r3}}
  & q \leftarrow \asum\lits{1:p} > 0
\end{alignat}
Considering the empty interpretation $\emptyset$,
we have that $\atoms{A}\cap X\neq \emptyset$ for each aggregate expression
$A\in\body{$\ref{eq1:r1}$}\cup\body{$\ref{eq1:r2}$}\cup\body{$\ref{eq1:r3}$}$.
Hence, $\tpu{\gz}{\prg_1}{X}{\infty}=\tpu{\gz}{\prg_1}{X}{0}=\emptyset$
yields that $X$ is not a \gz-answer set of~$\prg_1$.
Regarding $\tpx{\lpst}{\prg_1}{X}$,
observe that $\emptyset\nmodels\body{$\ref{eq1:r2}$}$,
$\emptyset\nmodels\body{$\ref{eq1:r3}$}$, and
$\{q\}\nmodels\body{$\ref{eq1:r1}$}$ in view of
$\eval{\asum\lits{1:p,-1:q} \geq 0}{\{q\}}=-1$.
This means that $(\emptyset,X)\unsat{\lpst} \body{r}$ for all $r\in\prg_1$,
so that $\tpu{\lpst}{\prg_1}{X}{\infty}=\tpu{\lpst}{\prg_1}{X}{0}=\emptyset$
disproves $X$ to be an \lpst-answer set of~$\prg_1$.

When we turn to $\tpx{\mr}{\prg_1}{X}$, then
$\eval{\asum\lits{1:p,-1:q} \geq 0}{\emptyset} =
 \eval{\asum\lits{1:p,-1:q} \geq 0}{X} = 0$
shows that $(\emptyset,X)\sat{\mr}\body{$\ref{eq1:r1}$}$ and
$\tpu{\mr}{\prg_1}{X}{1} = \tpc{\mr}{\prg_1}{X}{\emptyset} = \{p\}$.
Given that $(\{p\},X)\sat{\mr}\body{$\ref{eq1:r3}$}$,
this leads on to 
$\tpu{\mr}{\prg_1}{X}{\infty} =
 \tpu{\mr}{\prg_1}{X}{2} =
 \tpc{\mr}{\prg_1}{X}{\{p\}} = \{p,q\} = X$,
from which we conclude that $X$ is an \mr-answer set of $\prg_1$.

For obtaining the least fixpoint of $\tpx{\pdb}{\prg_1}{X}$,
consider the rules \ref{eq1:r1} and \ref{eq1:r2},
and observe that
$\emptyset\models \body{$\ref{eq1:r1}$}$,
$\{p\}\models \body{$\ref{eq1:r1}$}$,
$\{q\}\models \body{$\ref{eq1:r2}$}$,
and 
$X\models \body{$\ref{eq1:r1}$}$ as well as
$X\models \body{$\ref{eq1:r2}$}$.
That is, for each interpretation $Z\subseteq X$,
the body of some rule with $p$ in the head is satisfied by~$Z$.
Hence, we get
$\tpu{\pdb}{\prg_1}{X}{\nolinebreak 1} = \tpc{\pdb}{\prg_1}{X}{\emptyset} = \{p\}$,
which in view of $\{p\}\models \body{$\ref{eq1:r3}$}$ and
$X\models \body{$\ref{eq1:r3}$}$ brings us further to
$\tpu{\pdb}{\prg_1}{X}{\infty} = \tpu{\pdb}{\prg_1}{X}{2} =
 \tpc{\pdb}{\prg_1}{X}{\{p\}} = \{p,q\} = X$.
This means that $X$ is also a \pdb-answer set of $\prg_1$.

Let us modify the program $\prg_1$ to $\prg_1'$ over $\at=\{p,q,s\}$, in which
\ref{eq1:r1} and \ref{eq1:r3} are replaced by:\pagebreak[1]
\begin{alignat}{1}
  \label{eq1:r1p}\tag{$r_1'$}
  & s \leftarrow \asum\lits{1:p,-1:q} \geq 0
  \\
  \label{eq1:r3p}\tag{$r_3'$}  
  & q \leftarrow \asum\lits{1:s} > 0
\end{alignat}
Intuitively, rule $r_1'$ requires $s$ true if $p$ is true or $q$ is false, and rule $r_3'$ requires $q$ true if $s$ is true (recall that $r_2$ requires $p$ true if $q$ is true).
One can check that $X'=\{p,q,s\}$ is the unique model as well as an
\fflp- and \mr-answer set of $\prg_1'$.
Turning again to $\tpx{\pdb}{\prg_1'}{X'}$,
we have that
$\emptyset\nmodels \body{$\ref{eq1:r2}$}$,
$\emptyset\nmodels \body{$\ref{eq1:r3p}$}$,
$\{q\}\nmodels \body{$\ref{eq1:r1p}$}$,
$\{q\}\nmodels \body{$\ref{eq1:r3p}$}$,
and the heads of \ref{eq1:r1p} and \ref{eq1:r2}
are disjoint for the modified program $\prg_1'$.
(Each interpretation $Z'\subseteq X'$
 is such that $Z'\models \body{$\ref{eq1:r1p}$}$ or $Z'\models \body{$\ref{eq1:r2}$}$,
 yet $\head{$\ref{eq1:r1p}$}\cap \head{$\ref{eq1:r2}$} = \{s\} \cap \{p\} = \emptyset$.)
Given these considerations,
we conclude that
$\tpu{\pdb}{\prg_1'}{X'}{\infty} = \tpu{\pdb}{\prg_1'}{X'}{0} = \emptyset$,
so that $\prg_1'$ does not have any \pdb-answer set.
This shows that \fflp-answer sets are
not necessarily \pdb-answer sets as well.%
\end{example}

\begin{example}\label{ex:mr}
In Example~\ref{ex:subsum},
we have checked that $X=\{y_1,x_2,z_1,z_2,p\}$
is the unique \fflp-answer set of~$\prg_3$:\pagebreak[1]%
\begin{alignat}{1}
  \tag{\ref{eq6:r6}}
  & x_1 \leftarrow \asum\lits{1:y_1} < 1
  \\
  \tag{\ref{eq6:r7}}
  & y_1 \leftarrow \asum\lits{1:x_1} < 1
  \\
  \tag{\ref{eq6:r8}}
  & x_2 \leftarrow \asum\lits{1:y_2} < 1
  \\
  \tag{\ref{eq6:r9}}
  & y_2 \leftarrow \asum\lits{1:x_2} < 1
  \\
  \tag{\ref{eq6:r10}}
  & z_1 \leftarrow \asum\lits{1:p} > 0
  \\
  \tag{\ref{eq6:r11}}
  & z_2 \leftarrow \asum\lits{1:p} > 0
  \\
  \tag{\ref{eq6:r12}}
  & p \leftarrow \asum\lits{1:y_1,2:y_2,2:z_1,3:z_2} \neq 5
  \\
  \tag{\ref{eq6:r13}}
  & \bot \leftarrow \asum\lits{1:p} < 1
\end{alignat}
Inspecting $\tpx{\gz}{\prg_3}{X}$,
we get
$\tpu{\gz}{\prg_3}{X}{1} =
 \tpc{\gz}{\prg_3}{X}{\emptyset} = \{y_1,x_2\}$
because $X\models \body{$\ref{eq6:r7}$}$, $X\models \body{$\ref{eq6:r8}$}$, and
$\atoms{\asum\lits{1:x_1} < 1}\cap X = \atoms{\asum\lits{1:y_2} < 1}\cap X = \emptyset$.
Given that $X\nmodels \body{$\ref{eq6:r6}$}$, $X\nmodels \body{$\ref{eq6:r9}$}$,
$\atoms{\asum\lits{1:p} > 0}\cap \{y_1,x_2\} = \emptyset \neq \{p\} =
 \atoms{\asum\lits{1:p} > 0}\cap X$, and
$\atoms{\asum\lits{1:y_1,\linebreak[1]2:y_2,\linebreak[1]2:z_1,\linebreak[1]3:z_2} \neq 5}\cap \{y_1,x_2\} =
 \{y_1\} \neq \{y_1,z_1,z_2\} =
 \atoms{\asum\lits{1:y_1,2:y_2,2:z_1,3:z_2} \neq 5}\cap X$,
we obtain
$\tpu{\gz}{\prg_3}{X}{\infty} =
 \tpu{\gz}{\prg_3}{X}{1} = \{y_1,x_2\}$,
so that $X$ is not a \gz-answer set of $\prg_3$.

For $\any\in\{\lpst,\pdb\}$,
we also have that
$\tpu{\any}{\prg_3}{X}{1} =
 \tpc{\any}{\prg_3}{X}{\emptyset} = \{y_1,x_2\}$,
as $Z\models \body{$\ref{eq6:r7}$}$ and $Z\models \body{$\ref{eq6:r8}$}$
for each $Z\subseteq X$, while the interpretation
$\{z_1,z_2\}\subseteq X$ is such that
$\{z_1,z_2\}\nmodels \asum\lits{1:p}\linebreak[1] > 0$ and
$\{z_1,z_2\}\nmodels \asum\lits{1:y_1,2:y_2,2:z_1,3:z_2} \neq 5$.
However, $Z\models \asum\lits{1:y_1,\linebreak[1]2:y_2,\linebreak[1]2:z_1,\linebreak[1]3:z_2} \neq 5$ is the case
for every $\{y_1,x_2\}\subseteq Z\subseteq X$,
which leads to
$\tpu{\any}{\prg_3}{X}{2} =
 \tpc{\any}{\prg_3}{X}{\{y_1,x_2\}} = \{y_1,x_2,p\}$.
Then, $Z\models \asum\lits{1:p} > 0$, for any interpretation $\{y_1,x_2,p\}\subseteq Z$,
yields
$\tpu{\any}{\prg_3}{X}{\infty} =
 \tpu{\any}{\prg_3}{X}{3} =
 \tpc{\any}{\prg_3}{X}{\{y_1,x_2,p\}} = \{y_1,x_2,z_1,z_2,p\}=X$,
which shows that $X$ is an \lpst- and a \pdb-answer set of~$\prg_3$.

Regarding $\tpx{\mr}{\prg_3}{X}$,
in view of
$X\models \asum\lits{1:y_1,\linebreak[1]2:y_2,\linebreak[1]2:z_1,\linebreak[1]3:z_2} \neq 5$
and 
$\eval{\asum\lits{1:y_1,\linebreak[1]2:y_2,\linebreak[1]2:z_1,\linebreak[1]3:z_2} \neq 5}{\emptyset} = 0$,
we get
$\tpu{\mr}{\prg_3}{X}{1} =
 \tpc{\mr}{\prg_3}{X}{\emptyset} = \{y_1,x_2,p\}$ in the first step, and then
$\tpu{\mr}{\prg_3}{X}{\infty} =
 \tpu{\mr}{\prg_3}{X}{2} =
 \tpc{\mr}{\prg_3}{X}{\{y_1,x_2,p\}} = \{y_1,x_2,z_1,z_2,p\}=X$.
On the one hand, this means that $X$ is an \mr-answer set of $\prg_3$.
On the other hand,
the interpretations
$X_2=\{x_1,y_2,z_1,z_2,p\}$ and
$X_4=\{y_1,y_2,z_1,z_2,p\}$ (according to the naming scheme of Example~\ref{ex:subsum})
are obtained as additional \mr-answer sets of $\prg_3$ that are
not backed up by any of the other semantics.
In fact,
counting on the existence of some interpretation $Z \subseteq \tpu{\mr}{\prg}{X}{i}$
such that $Z\models A$, for an aggregate expression~$A$ belonging to the body of a rule $r\in \prg$,
to proceed with (potentially) adding $\head{r}$ to $\tpu{\mr}{\prg}{X}{i}$
disregards the satisfaction of $A$ by $\tpu{\mr}{\prg}{X}{i}$ itself as well as interpretations
extending it.
\begin{rev}
For this reason, \citeN{DBLP:journals/corr/abs-2104-14789}
classify $\tpx{\mr}{\prg}{X}$ as not well-behaved,
while $\tpx{\any}{\prg}{X}$
is well-behaved for $\any\in\{\gz,\lpst,\pdb\}$.
\end{rev}
\end{example}

The programs considered in Example \ref{ex:pdb}--\ref{ex:mr} did not have
\gz-answer sets because of the strong condition
$\atoms{A}\cap Y = \atoms{A}\cap X$ for $(Y,X)\sat{\gz} A$.
Its intention is to circumvent so-called vicious circles \cite{gelzha19a},
where the satisfaction of an aggregate expression~$A$ has some influence on the
outcome of the aggregation function in~$A$,
which happens in Example \ref{ex:pdb}--\ref{ex:mr}, e.g.,
with the rule $p \leftarrow \asum\lits{1:p,-1:q} \geq 0$
to conclude~$p$ from an aggregate expression mentioning~$p$.

To make circularity more formal, for a program~\prg, let
$\gp{\prg}=(\at,\{(p,q) \mid r\in\prg,\linebreak[1] p\in\head{r},\linebreak[1] A\in\body{r},\linebreak[1] q\in\atoms{A}\})$
be the (directed) \emph{atom dependency graph} of~\prg.
If $\gp{\prg}$ is acyclic (and the rules in \prg are non-disjunctive),
one can show that there is at most one \any-answer set of \prg,
for $\any\in\{\fflp,\gz,\lpst,\mr,\pdb\}$,
and that all aggregate semantics under consideration coincide.
Provided that \prg does not include constraints,
acyclicity of $\gp{\prg}$ is sufficient for the existence
of a \gz-answer set of \prg.
However, such acyclicity is not a necessary condition, e.g.,
\gz-answer sets match standard answer sets \cite{gellif88b}
when aggregate expressions of the form
$\asum\lits{1:p}>0$ or $\asum\lits{1:p}<1$
replace (negated) propositional atoms~$p$,
no matter whether $\gp{\prg}$ is acyclic or not,
and this correspondence applies to all but the \pdb-answer set semantics
(cf.\ Example~\ref{ex:pdb}).
In fact, vicious circles are not identified syntactically,
but rather the construction of a \gz-answer set $X$ of \prg
by $\tpx{\gz}{\prg}{X}$ witnesses the non-circular provability of all atoms in~$X$.

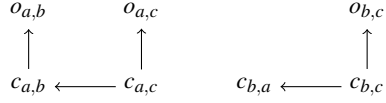
\begin{figure}
    \figrule
    \centering

    \tikzstyle{none} = [text centered]
    \tikzstyle{line} = [draw, ->]
       
    \begin{tikzpicture}
        \node at (0,0) [none](oab) {$o_{a,b}$};
        \node at (1.5,0) [none](oac) {$o_{a,c}$};
        \node at (4.5,0) [none](obc) {$o_{b,c}$};
        \node at (0,-1) [none](cab) {$c_{a,b}$};
        \node at (1.5,-1) [none](cac) {$c_{a,c}$};
        \node at (3,-1) [none](cba) {$c_{b,a}$};
        \node at (4.5,-1) [none](cbc) {$c_{b,c}$};
        
        \path[line] (cab) -- (oab);
        \path[line] (cac) -- (oac);
        \path[line] (cac) -- (cab);
        \path[line] (cbc) -- (obc);
        \path[line] (cbc) -- (cba);
    \end{tikzpicture}
    \caption{Dependency graph of program $\prg_4$ from Example~\ref{ex:gz}.}\label{fig:company-depgraph}
    \figrule
\end{figure}

\begin{example}\label{ex:gz}
Consider the program $\prg_4$,
which constitutes a propositional version of the company controls problem encoding
introduced in Section~\ref{sec:introduction} for the instance given in Figure~\ref{fig:company}:\pagebreak[1]
\begin{alignat}{1}
  \label{eq11:r14}\tag{$r_{14}$}
  & o_{a,b} \leftarrow \top
  \\
  \label{eq11:r15}\tag{$r_{15}$}
  & o_{a,c} \leftarrow \top
  \\
  \label{eq11:r16}\tag{$r_{16}$}
  & o_{b,c} \leftarrow \top
  \\
  \label{eq11:r17}\tag{$r_{17}$}
  & c_{a,b} \leftarrow \asum\lits{80: o_{a,b}} > 50
  \\
  \label{eq11:r18}\tag{$r_{18}$}
  & c_{a,c} \leftarrow \asum\lits{30: o_{a,c}, 30: c_{a,b}} > 50
  \\
  \label{eq11:r19}\tag{$r_{19}$}
  & c_{b,c} \leftarrow \asum\lits{30: o_{b,c}, 30: c_{b,a}} > 50
\end{alignat}
The atom dependency graph
$\gp{\prg_4}=(\{o_{a,b},o_{a,c},o_{b,c},c_{a,b},c_{a,c},c_{b,a},c_{b,c}\},
              \{(c_{a,b},o_{a,b}),\linebreak[1]
                (c_{a,c},o_{a,c}),\linebreak[1](c_{a,c},c_{a,b}),\linebreak[1]
                (c_{b,c},o_{b,c}),\linebreak[1](c_{b,c},c_{b,a})\})$,
shown in Figure~\ref{fig:company-depgraph},
is acyclic, and the interpretation $X=\{o_{a,b},o_{a,c},o_{b,c},c_{a,b},c_{a,c}\}$
is a model of $\prg_4$.
Let us construct $\tpu{\gz}{\prg_4}{X}{\infty}$ to check that
$X$ is a \gz-answer set of $\prg_4$.
In the first step, $(\emptyset,X)\sat{\gz}\body{r}$ applies for the facts
$r\in\{$\ref{eq11:r14}$,$\ref{eq11:r15}$,$\ref{eq11:r16}$\}$, so that
$\tpu{\gz}{\prg_4}{X}{1} =
 \tpc{\gz}{\prg_4}{X}{\emptyset} =
 \{o_{a,b},\linebreak[1]o_{a,c},\linebreak[1]o_{b,c}\}$.
Then, we have that
$\atoms{\asum\lits{80: o_{a,b}} > 50}\cap \tpu{\gz}{\prg_4}{X}{1} =
 \atoms{\asum\lits{80: o_{a,b}} > 50}\cap X = \{o_{a,b}\}$, and
$X\models \body{$\ref{eq11:r17}$}$ in view of
$\eval{\asum\lits{80: o_{a,b}} > 50}{X}=80$.
We thus get
$\tpu{\gz}{\prg_4}{X}{2} =
 \tpc{\gz}{\prg_4}{X}{\{o_{a,b},\linebreak[1]o_{a,c},\linebreak[1]o_{b,c}\}} =
 \{o_{a,b},\linebreak[1]o_{a,c},\linebreak[1]o_{b,c},\linebreak[1]c_{a,b}\}$,
and
$\atoms{\asum\lits{30: o_{a,c}, 30: c_{a,b}} > 50}\cap \tpu{\gz}{\prg_4}{X}{\nolinebreak 2} =
 \atoms{\asum\lits{30: o_{a,c}, 30: c_{a,b}} > 50}\cap X = \{o_{a,c},\linebreak[1]c_{a,b}\}$
along with
$\eval{\asum\lits{30: o_{a,c},\linebreak[1] 30: c_{a,b}} > 50}{X}=60$ and
$X\models \body{$\ref{eq11:r18}$}$ lead to
$\tpu{\gz}{\prg_4}{X}{3} =
 \tpc{\gz}{\prg_4}{X}{\{o_{a,b},\linebreak[1]o_{a,c},\linebreak[1]o_{b,c},\linebreak[1]c_{a,b}\}} =
 \{o_{a,b},\linebreak[1]o_{a,c},\linebreak[1]o_{b,c},\linebreak[1]c_{a,b},\linebreak[1]c_{a,c}\} = X$.
Given that $X\nmodels \body{$\ref{eq11:r19}$}$, the least fixpoint
$\tpu{\gz}{\prg_4}{X}{\infty} = \tpu{\gz}{\prg_4}{X}{3} = X$
yields that $X$ is a \gz-answer set of $\prg_4$.

However, note that $X$ is no longer a \gz-answer set of $\prg_4'$,
obtained by replacing \ref{eq11:r17} with
\begin{alignat}{1}
  \label{eq11:r17p}\tag{$r_{17}'$}
  & c_{a,b} \leftarrow \asum\lits{80: o_{a,b}, w: c_{a,c}} > 50
\end{alignat}
for some weight $w\geq 0$.
While we still obtain
$\tpu{\gz}{\prg_4'}{X}{1} =
 \tpc{\gz}{\prg_4'}{X}{\emptyset} =
 \{o_{a,b},\linebreak[1]o_{a,c},\linebreak[1]o_{b,c}\}$,
we end up with
$\tpu{\gz}{\prg_4'}{X}{\infty} =
 \tpu{\gz}{\prg_4'}{X}{1}$
because
$\atoms{\asum\lits{80: o_{a,b}, w: c_{a,c}} > 50}\cap \tpu{\gz}{\prg_4'}{X}{1}
 = \{o_{a,b}\} \neq \{o_{a,b},\linebreak[1]c_{a,c}\}=
 \atoms{\asum\lits{80: o_{a,b}, w: c_{a,c}} > 50}\cap X$
and thus
$(\tpu{\gz}{\prg_4'}{X}{1},X)\unsat{\gz} \body{$\ref{eq11:r17p}$}$.
That is, $X$ is rejected as a $\gz$-answer set of $\prg_4'$ in view of
the circularity between $c_{a,b}$ and $c_{a,c}$,
which are
involved in aggregate expressions in the bodies of \ref{eq11:r17p} and \ref{eq11:r18}.
When considering standard answer sets of programs without (sophisticated) aggregates
\cite{gellif88b,gellif91a},
circular rules alone do not yield provability of atoms in their heads.
Unlike that,
not all propositional atoms subject to an aggregation function may be needed
to satisfy a respective aggregate expression,
so that rejecting any circularity is a very strong condition for programs with aggregates.
In fact, without going into the details,
we have that $X$ is the unique \any-answer set of both $\prg_4$ and $\prg_4'$
for $\any\in\{\fflp,\lpst,\mr,\pdb\}$, i.e.,
all aggregate semantics other than \gz.%
\end{example}

\subsection{Semantic Relationships}\label{subsec:relationships}

As mentioned above, the satisfaction relation $\sat{\gz}$ is a subset of
$\sat{\lpst}$, which is in turn a subset of $\sat{\mr}$.
That is, given a model $X$ of a (non-disjunctive) program~\prg,
we immediately get
$\tpu{\lpst}{\prg}{X}{\infty} = X$ from
$\tpu{\gz}{\prg}{X}{\infty} = X$, and
$\tpu{\mr}{\prg}{X}{\infty} = X$ from
$\tpu{\lpst}{\prg}{X}{\infty} = X$,
which means that a \gz-answer set $X$ of \prg is an \lpst-answer set as well,
and an \lpst-answer set $X$ of \prg also an \mr-answer set.
The converse relationships do not apply in general, yet may come into effect for
programs restricted to monotone, anti-monotone, or convex aggregate expressions only.
Moreover, the question arises how \pdb-answer sets determined by the operator
$\tpx{\pdb}{\prg}{X}$ and model-based \fflp-answer sets relate to the other
aggregate semantics.
A part of these relationships have already been studied by
\citeN{ferraris11a} and \citeN{liposotr10a},
and in the following we give a complete account for the aggregate semantics
under consideration, assuming programs in question to be non-disjunctive.

Figure~\ref{fig:answersets} \begin{rev}and Table~\ref{tab:answersets} visualize\end{rev} the relationships between
different aggregate semantics, where arcs without label express that
answer sets under one semantics are preserved by another semantics for
programs with arbitrary, i.e., possibly non-convex, aggregate expressions.
The restriction of such a correspondence to programs with convex, monotone,
or anti-monotone aggregate expressions only is indicated by the arc label $\pm$,
$+$, or $-$, respectively.
Transitive relationships, e.g., the fact that each \gz-answer set is an
\mr-answer set as well, are not stated by explicit arcs for better readability.
A summary of the logic programs used to
illustrate the similarities and differences between the aggregate semantics 
in \begin{rev}Sections~\ref{subsec:model} and~\ref{subsec:construction}\end{rev}
is given in Table~\ref{tab:relationships}. 
%
\begin{figure}[t]
\figrule
\centering
        \tikzstyle{none} = [text centered]
        \tikzstyle{line} = [draw, ->]
        \begin{tikzpicture}
            \node at (0.5,0)    [none](gz)   {\gz};
            \node at (3,0)    [none](lpst) {\lpst};
            \node at (6,0)    [none](pdb)  {\pdb};
            \node at (3,-2) [none](fflp) {\fflp};
            \node at (5.5,-2) [none](mr)   {\mr};
            
            \path[line] (gz) -- (lpst);
            \path[line] (lpst) -- (pdb);
            \path[line] (pdb) to[bend right=20] node[midway, above] {$+$} (lpst);
            \path[line] (pdb) to[bend left=20] node[midway, below] {$-$} (lpst);
            \path[line] (lpst) to (fflp);
            \path[line] (fflp) to (mr);
            \path[line] (mr) to node[midway, below] {$\pm$} (lpst);
        \end{tikzpicture}
\capvspace
\caption{Relationships between aggregate semantics,
         where an arc indicates that answer sets carry forward from
         the origin to the target semantics for non-disjunctive programs with
         monotone (arc label $+$),
         anti-monotone (arc label $-$),
         convex (arc label $\pm$), or
         arbitrary (no arc label) aggregate expressions, respectively.\label{fig:answersets}}
\figrule
\end{figure}
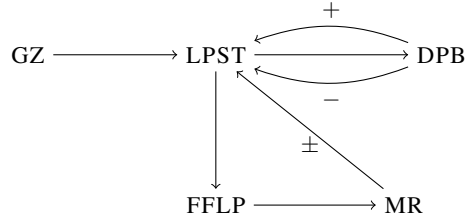
\begin{table}[t]
    \caption{
\protect\begin{rev}%
          Monotonicity conditions guaranteeing answer sets according to the semantics in rows to be answer sets according to the semantics in columns.
\protect\end{rev}%
    }\label{tab:answersets}
\begin{rev}
    \begin{tabular}{cccccc}
        \hline\hline
        & \gz & \lpst & \pdb & \fflp & \mr \\
        \hline
        \gz & --- & arbitrary & arbitrary & arbitrary & arbitrary \\
        \lpst & none & --- & arbitrary & arbitrary & arbitrary \\
        \pdb & none & (anti-)monotone & --- & (anti-)monotone & (anti-)monotone \\
        \fflp & none & convex & convex & --- & arbitrary \\
        \mr & none & convex & convex & convex & ---\\
        \hline\hline
    \end{tabular}
\end{rev}
\end{table}
\begin{table}[t]
    \caption{Summary of examples showing differences between aggregate semantics, where $X = \{o_{a,b},o_{a,c},o_{b,c},c_{a,b},c_{a,c}\}$.}\label{tab:relationships}
    \begin{tabular}{cccccc}
        \hline\hline
        Program & \fflp & \gz & \lpst & \mr & \pdb \\
        \hline
        $\prg_1$ & $\{p,q\}$ & --- & --- & $\{p,q\}$ & $\{p,q\}$ \\
        Example~\ref{ex:model} & Example~\ref{ex:minmodel} & Example~\ref{ex:fflp} & Example~\ref{ex:fflp} & Example~\ref{ex:fflp} & Example~\ref{ex:fflp} \\ 
        \,\\
        $\prg_1'$ & $\{p,q,s\}$ & --- & --- & $\{p,q,s\}$ & --- \\
        Example~\ref{ex:fflp} & Example~\ref{ex:fflp} & Example~\ref{ex:fflp} & Example~\ref{ex:fflp} & Example~\ref{ex:fflp} & Example~\ref{ex:fflp} \\
        \,\\
        $\prg_2$ & --- & --- & --- & --- & $\{p\}$ \\
        Example~\ref{ex:unsatisfiable} & Example~\ref{ex:unsatisfiable} & Example~\ref{ex:pdb} & Example~\ref{ex:pdb} & Example~\ref{ex:pdb} & Example~\ref{ex:pdb} \\
        \,\\
        & & & & $\{x_1,y_2,z_1,z_2,p\}$, \\
        & & & & $\{y_1,y_2,z_1,z_2,p\}$, \\
        $\prg_3$ & $\{y_1,x_2,z_1,z_2,p\}$ & --- & $\{y_1,x_2,z_1,z_2,p\}$ & $\{y_1,x_2,z_1,z_2,p\}$\, & $\{y_1,x_2,z_1,z_2,p\}$ \\
        Example~\ref{ex:subsum} & Example~\ref{ex:subsum} & Example~\ref{ex:mr} & Example~\ref{ex:mr} & Example~\ref{ex:mr} & Example~\ref{ex:mr} \\
        \,\\
        $\prg_4$ & $X$ & $X$ & $X$ & $X$ & $X$ \\
        Example~\ref{ex:gz} & Example~\ref{ex:gz} & Example~\ref{ex:gz} & Example~\ref{ex:gz} & Example~\ref{ex:gz} & Example~\ref{ex:gz} \\
        \,\\
        $\prg_4'$ & $X$ & --- & $X$ & $X$ & $X$ \\
        Example~\ref{ex:gz} & Example~\ref{ex:gz} & Example~\ref{ex:gz} & Example~\ref{ex:gz} & Example~\ref{ex:gz} & Example~\ref{ex:gz} \\
        \hline\hline
    \end{tabular}
\end{table}

The most restrictive semantics is obtained with \gz-answer sets, in the sense
that the fewest models $X$ of a program $\prg$ pass the construction by means of $\tpx{\gz}{\prg}{X}$.
As $\tpx{\lpst}{\prg}{X}$ relaxes the conditions for concluding head atoms,
\lpst-answer sets are guaranteed to include \gz-answer sets, as indicated by
an arc without label in Figure~\ref{fig:answersets}.
The observation that $(Y,X)\sat{\lpst} \body{r}$, for some rule $r\in\prg$ and $Y\subseteq X$,
implies $Z\models \body{r}$ for every interpretation $Y\subseteq Z\subseteq X$
yields that $\head{r}\subseteq \tpc{\pdb}{\prg}{X}{Y}$,
so that each \lpst-answer set $X$ of \prg is a \pdb-answer set as well.
Moreover, when $Y\subset X$ for an \lpst-answer set $X$ of \prg,
the fact that $\tpc{\lpst}{\prg}{X}{Y}\nsubseteq Y$ shows that $Y$ is not a model
of the reduct $\reduct{\prg}{X}$,
which in turn witnesses that $X$ is a $\subseteq$-minimal model of $\reduct{\prg}{X}$
and thus an \fflp-answer set of \prg.
The last general relationship that applies for programs with arbitrary aggregate expressions
expresses that each \fflp-answer set~$X$ of \prg is also an \mr-answer set, given that
$Y\nmodels \reduct{\prg}{X}$, for any interpretation $Y\subset X$,
implies that $\tpc{\mr}{\prg}{X}{Y}\nsubseteq Y$.
As discussed in \begin{rev}Examples~\ref{ex:pdb} and~\ref{ex:fflp}\end{rev},
\fflp- and \pdb-answer sets can be mutually distinct,
so that no (transitive) general relationship is obtained between these two aggregate semantics.

Considering relationships for convex, monotone, or anti-monotone aggregate expressions,
an \mr-answer set $X$ of \prg is also an \lpst-answer set when the program \prg
includes convex aggregate expressions only.
This follows from the observation that, for 
a convex aggregate expression $A$ and each interpretation $Y\subseteq X$,
the conditions $X\models A$ and $Z\models A$, for some $Z\subseteq Y$, of
$(Y,X)\sat{\mr} A$ yield the condition of $(Y,X)\sat{\lpst} A$ that
$Z\models A$ for every $Y\subseteq Z\subseteq X$.
The general relationship between \lpst- and \fflp-answer sets then further
implies that $X$ is an \fflp-answer set of~\prg as well,
so that \fflp-, \lpst-, and \mr-answer sets coincide for programs with
convex aggregate expressions only.
For establishing similar correspondences to \pdb-answer sets,
as Example~\ref{ex:pdb} shows,
programs with convex or both monotone and anti-monotone
aggregate expressions, respectively, are still too general.
However, when the aggregate expressions in a program \prg are monotone or anti-monotone only,
for interpretations $Y\subseteq X$,
the intersection over all $Y\subseteq Z\subseteq X$ taken by
$\tpc{\pdb}{\prg}{X}{Y}$ collapses to checking whether rule bodies
are satisfied by $Y$ or $X$, respectively,
while satisfaction by other interpretations $Z$ is a consequence of
(anti-)monotonicity.
Hence, we have that $\tpc{\lpst}{\prg}{X}{Y}=\tpc{\pdb}{\prg}{X}{Y}$,
which means that each \pdb-answer set~$X$ of \prg is also an \lpst-answer set
in case of monotone or anti-monotone aggregate expressions only.
By (transitive) general relationships,
this correspondence carries forward to \fflp- and \mr-answer sets.

\begin{rev}
Interestingly, \gz-answer sets do not correspond to the other semantics even for programs whose aggregate expressions are restricted to be monotone.
\end{rev}
For instance,
the program \prg consisting of the rule
$p \leftarrow \asum\lits{1:p} \geq 0$,
which includes the monotone as well as anti-monotone aggregate expression $\asum\lits{1:p} \geq 0$,
has $\{p\}$ as an answer set under all aggregate semantics but \gz,
where the latter is due to 
$\tpu{\gz}{\prg}{\{p\}}{\infty}=\tpu{\gz}{\prg}{\{p\}}{0}=\emptyset$
in view of $\atoms{\asum\lits{1:p} \geq 0}\cap \{p\}\neq \emptyset$.
As $\{p\}$ is the unique model of \prg,
this means that \prg does not have any \gz-answer set,
even though \prg includes only one aggregate expression that is both monotone and anti-monotone.
Such a behavior is justified by \citeN{gelzha19a} by a more uniform treatment of recursive symbolic rules like the following:
\begin{asp}
  p :- #sum{1 : p} >= 0.
  p :- #sum{1 : p} = S, S >= 0.
\end{asp}
In fact, a program consisting of the second rule above, would be instantiated as\pagebreak[1]
\begin{alignat*}{1}
  & p \leftarrow \asum\lits{1: p} = 0\\
  & p \leftarrow \asum\lits{1: p} = 1  
\end{alignat*}
which has no answer sets according to all semantics discussed in this paper.


\section{Computational Complexity of Aggregates}\label{sec:complexity}

The expressiveness of programs with aggregate expressions can be assessed in terms of the
computational complexity \begin{rev}\cite{garjoh79}\end{rev} of specific reasoning tasks.
To this end, we investigate the following two decision problems for $\any\in\{\fflp,\gz,\lpst,\mr,\pdb\}$:
\begin{description}
\item[\textnormal{\checkas:}]
For a program \prg and an interpretation $X\subseteq\at$,
decide whether $X$ is a \any-answer set of~\prg.
\item[\textnormal{\existas:}]
For a program \prg, decide whether there is some \any-answer set of~\prg.
\end{description}
The complexity of both tasks is well-understood \begin{rev}\cite{daeigovo01a}\end{rev}
for programs \prg in which rule bodies consist of (negated) propositional atoms~$p$,
corresponding to aggregate expressions of the form $\asum\lits{1:p}>0$ or $\asum\lits{1:p}<1$,
under standard answer set semantics \cite{gellif88b,gellif91a}.
Here the complexity of reasoning tasks depends on whether \prg is non-disjunctive or not,
\checkas is \poly-complete (i.e., tractable) in the non-disjunctive case,
and \coNP-complete otherwise.
The \existas task, where no interpretation $X\subseteq \at$ is fixed,
is \NP-complete when \prg is non-disjunctive,
and in general \stwo-complete in the presence of (proper) disjunctive rules. 
The latter means that two orthogonal combinatorial problems are interleaved,
one about determining candidate models~$X$ among an exponential number of interpretations,
and the other concerning the check that no $Y\subset X$ is a model of the reduct relative to~$X$ (again, the search space for $Y$ is exponential).

Given that (proper) disjunctive rules lead to elevated computational complexity
already for the simple forms of aggregate expressions resembling (negated) propositional atoms,
we again assume programs to be non-disjunctive in the following.
This allows us to study the complexity of the \checkas and \existas tasks
for \any-answer sets relative to $\any\in\{\fflp,\gz,\lpst,\mr,\pdb\}$
and the monotonicity of aggregate expressions,
where the obtained complexity classes are summarized in Table~\ref{tab:complexity}.
Except for the complexity class~\poly, for which we indicate membership but not
necessarily hardness, our considerations address completeness, i.e.,
membership along with hardness for some complexity class beyond~\poly.
In Section~\ref{subsec:check},
we investigate the computational complexity of the \checkas task in detail,
and Section~\ref{subsec:exist} provides a closer study of the \existas task.
\begin{table}[t]
\caption{Computational complexity of the \existas and \checkas
         reasoning tasks for different aggregate semantics and the
         monotonicity of aggregate expressions included in non-disjunctive programs,
         where cases in which elevated complexity relies on the \NP-hardness of
         deciding whether aggregate expressions under consideration are satisfiable
         are indicated by lower complexity classes obtained with tractable 
         satisfiability checks in parentheses.\label{tab:complexity}}
\setlength{\tabespace}{\widthof{)}}
\tabrspace
\begin{tabular}{rrrrrrl}
\hline\hline
Monotonicity  & \gz & \mr & \lpst & \fflp & \pdb &
\\
\hline
monotone      & \poly & \poly                 & \poly                   & \poly & \poly & \multirow{4}{*}{\rotatebox{270}{\checkas\tabhspace}}
\\
anti-monotone & \poly & \poly                 & \poly                   & \poly & \poly & 
\\
convex        & \poly & \poly                 & \poly                   & \poly & \coNP & 
\\
arbitrary     & \poly & \NP (\poly)\tabbspace & \coNP (\poly)\tabbspace & \coNP & \coNP & 
\\
\hline
monotone      & \poly & \poly                 & \poly                   & \poly & \poly & \multirow{4}{*}{\rotatebox{270}{\existas\tabhspace}}
\\
anti-monotone & \NP   & \NP                   & \NP                     & \NP   & \NP   & 
\\
convex        & \NP   & \NP                   & \NP                     & \NP   & \stwo & 
\\
arbitrary     & \NP   & \NP                   & \stwo (\NP)\tabbspace   & \stwo & \stwo & 
\\
\hline\hline
\end{tabular}
\end{table}
%

\subsection{Answer Set Checking}\label{subsec:check}

Starting with \checkas for (non-disjunctive) programs \prg including monotone or
anti-monotone aggregate expressions only,
we have that the task stays in \poly for all \begin{rev}aggregate\end{rev} semantics.
An intuitive explanation is that all but the \gz-answer set semantics coincide
for such programs and that a construction, e.g., by means of $\tpc{\pdb}{\prg}{X}{Y}$,
for some $Y\subseteq X$, can focus on concluding head atoms of rules $r\in\prg$
such that $Y\models \body{r}$ or $X\models \body{r}$, respectively.
\begin{rev}For $\tpc{\gz}{\prg}{X}{Y}$, also the additional condition $\atoms{A}\cap Y = \atoms{A}\cap X$
needs to be checked for each aggregate expression $A\in\body{r}$.
Regardless of the particular semantics, all relevant checks can be accomplished in polynomial time:
$X\models \prg$ to make sure that $X$ is a model of~\prg,
$Y\models \body{r}$ or $X\models \body{r}$ for rules $r\in\prg$, and possibly
$\atoms{A}\cap Y = \atoms{A}\cap X$ for aggregate expressions $A\in\body{r}$.\end{rev}
The strong condition $\atoms{A}\cap Y = \atoms{A}\cap X$ for $(Y,X)\sat{\gz} A$
also circumvents elevated complexity of \checkas under \gz-answer set semantics
when turning to programs with convex or arbitrary, i.e., possibly non-convex, aggregate expressions.

In case of convex aggregate expressions, which include monotone as well as anti-monotone
aggregate expressions,
the \checkas task remains tractable for $\any\in\{\fflp,\lpst,\mr\}$,
given that the three semantics coincide, and for an aggregate expression~$A$, checking 
$Y\models A$ and $X\models A$ is sufficient to conclude that
$(Y,X)\sat{\lpst} A$ as well as $(Y,X)\sat{\mr} A$.
Unlike that, \checkas becomes more complex, i.e., \coNP-complete,
for \pdb-answer sets of programs with convex aggregate expressions, where
\citeN{dematr04a} show \coNP-hardness by reduction from deciding whether
a propositional formula in disjunctive normal form is a tautology.
Notably, the logic program given in this reduction does not refer to
(sophisticated) aggregates, and the simultaneous availability of
monotone and anti-monotone
aggregate expressions of the form $\asum\lits{1:p}>0$ or $\asum\lits{1:p}<1$
to express (negated) propositional atoms~$p$ is already sufficient.
This complexity does also not increase any further in the presence of
non-convex aggregate expressions, as for disproving $X$ to be a
\pdb-answer set of \prg, it is sufficient to check that $X\nmodels \prg$,
or otherwise to determine a collection of (not necessarily distinct) interpretations
$Y_1\subseteq X,\dots,Y_{|X|}\subseteq X$ such that
$Z\subseteq Y$ and $Z\subset X$ hold for
$Y= Y_1\cap \dots\cap Y_{|X|}$ and
\begin{alignat*}{1}
  Z=\left( \mbox{$\bigcup$}_{r\in\prg,Y_1    \models\body{r}}\head{r} \right)\cap \dots\cap
   \left( \mbox{$\bigcup$}_{r\in\prg,Y_{|X|}\models\body{r}}\head{r} \right)\cap X\text{.}
\end{alignat*}
In the latter case, we have that
$\tpc{\pdb}{\prg}{X}{Y}\subseteq Y$ and
$\tpc{\pdb}{\prg}{X}{Y}\subset X$ for
$Y = Y_1\cap \dots\cap Y_{|X|}\subseteq X$,
which in turn implies that $\tpu{\pdb}{\prg}{X}{\infty}\subseteq\tpc{\pdb}{\prg}{X}{Y}\subset X$.
Moreover, when $\tpu{\pdb}{\prg}{X}{\infty}\subset X$, note that
at most $|X\setminus\tpu{\pdb}{\prg}{X}{\infty}|\leq |X|$ many 
interpretations $\tpu{\pdb}{\prg}{X}{\infty}\subseteq Y\subseteq X$
are needed to show that none of the atoms in $X\setminus\tpu{\pdb}{\prg}{X}{\infty}$
belongs to $\tpc{\pdb}{\prg}{X}{\tpu{\pdb}{\prg}{X}{\infty}}$,
so that \checkas under \pdb-answer set semantics stays in \coNP
for programs with arbitrary aggregate expressions.

For $\any\in\{\fflp,\lpst,\mr\}$,
arbitrary aggregate expressions make a difference regarding the complexity of the \checkas task.
Concerning \fflp-answer sets, membership in \coNP is immediate as checking that
$X\nmodels \prg$ or $Y\models \reduct{\prg}{X}$, for some $Y\subset X$, is tractable.
The \coNP-hardness can be established by a reduction from disjunctive programs to
programs with nested implications due to \citeN{ferraris11a}, which replaces a
disjunctive rule like $p\vee q\leftarrow \top$ by two rules
$p\leftarrow \underline{(p\leftarrow q)}$ and $q\leftarrow \underline{(q\leftarrow p)}$.
(Note that rule bodies with nested implications are underlined to avoid any ambiguity with rules.)
These nested implications are satisfied by the empty interpretation~$\emptyset$,
become unsatisfied by $\{q\}$ or $\{p\}$, respectively, and are satisfied by
the remaining interpretations, in particular, the extended interpretation $\{p,q\}$.
Such satisfaction can in turn be captured in terms of non-convex aggregate expressions like
$\asum\lits{1:p,-1:q}\geq 0$ and $\asum\lits{-1:p,1:q}\geq 0$ for $\underline{(p\leftarrow q)}$
or $\underline{(q\leftarrow p)}$, respectively, and various other representations, e.g.,
$\amax\lits{1:p,-1:q}\neq -1$ or $\aavg\lits{0:x,-1:p,1:q}\geq 0$, where $x\leftarrow \top$
is used as an auxiliary fact, yield the same effect.
This shows that \checkas under \fflp-answer set semantics is \coNP-complete for programs
including non-convex aggregate expressions \begin{rev}\cite{alvfab13a}\end{rev}.

Turning to \lpst-answer sets, an interpretation $Y\subset X$ disproves a model $X$
of \prg to be an \lpst-answer set of \prg, if for each rule $r\in\prg$
with $\head{r}\nsubseteq Y$,
we find an aggregate expression $A\in\body{r}$ along with some $Y\subseteq Z\subseteq X$
such that $Z\nmodels A$.
As the problem of checking the existence of such a counterexample $Y$ belongs to \NP,
the complementary \checkas task for \lpst-answer sets is a member of \coNP.
The \coNP-hardness though depends on the complexity of deciding whether
$(Y,X)\sat{\lpst} A$ for an aggregate expression~$A$,
which is intractable \ifaoif deciding the satisfiability
of the complement of~$A$, obtainable by inverting the comparison operator of~$A$,
e.g.,
$\agg\lits{w_1:p_1,\dots,w_n:p_n}=\bound$
when $A$ is of the form $\agg\lits{w_1:p_1,\dots,w_n:p_n}\neq\bound$,
is \NP-complete.
As discussed in Section~\ref{subsec:satisfiability},
such elevated complexity applies to $\agg\in\{\asum,\linebreak[1]\atimes,\linebreak[1]\aavg\}$
in combination with the (complementary) comparison operator~$=$.
Hence, \checkas for \lpst-answer sets is only \coNP-complete
when \prg includes aggregate expressions
$\agg\lits{w_1:p_1,\linebreak[1]\dots,\linebreak[1]w_n:p_n}\neq\bound$
with $\agg\in\{\asum,\linebreak[1]\atimes,\linebreak[1]\aavg\}$,
while it drops to \poly otherwise,
even in the presence of other forms of non-convex aggregate expressions like
$\aavg\lits{w_1:p_1,\linebreak[1]\dots,\linebreak[1]w_n:p_n}\geq\bound$ or
$\amin\lits{w_1:p_1,\linebreak[1]\dots,\linebreak[1]w_n:p_n}\neq\bound$.

\begin{example}\label{ex:checklpst}
Consider the program~$\prg_5$ as follows:\pagebreak[1]
\begin{alignat}{1}
  \label{eq12:r20}\tag{$r_{20}$}
  & x_1 \leftarrow \asum\lits{1:p} > 0
  \\
  \label{eq12:r21}\tag{$r_{21}$}
  & x_2 \leftarrow \asum\lits{1:p} > 0
  \\
  \label{eq12:r22}\tag{$r_{22}$}
  & x_3 \leftarrow \asum\lits{1:p} > 0
  \\
  \label{eq12:r23}\tag{$r_{23}$}
  & p \leftarrow \asum\lits{2 : x_1, 2 : x_2, 3 : x_3} \neq 5
\end{alignat}
For deciding whether the interpretation $X=\{x_1,x_2,x_3,p\}$
is an \lpst-answer set of $\prg_5$,
starting from $\tpu{\lpst}{\prg_5}{X}{0}=\emptyset$,
we need to check the condition of $(\emptyset,X)\sat{\lpst} \body{$\ref{eq12:r23}$}$
that $Z\models \asum\lits{2 : x_1,\linebreak[1] 2 : x_2,\linebreak[1] 3 : x_3} \neq 5$ for every $Z\subseteq X$.
This condition fails \ifaoif the complementary aggregate expression
$A = \asum\lits{2 : x_1,\linebreak[1] 2 : x_2,\linebreak[1] 3 : x_3} = 5$
is satisfiable, i.e.,
when $\eval{A}{Z} = 5$ for some $Z\subseteq X$.
As $Z_1=\{x_1,x_3\}$ and $Z_2=\{x_2,x_3\}$ are such that
$\eval{A}{Z_1} = \eval{A}{Z_2} = 5$, the complementary aggregate expression~$A$
is satisfiable, so that $(\emptyset,X)\unsat{\lpst} \body{$\ref{eq12:r23}$}$ and
$\tpu{\lpst}{\prg_5}{X}{\infty}=\tpu{\lpst}{\prg_5}{X}{0}=\emptyset$ disproves $X$
to be an \lpst-answer set of $\prg_5$.

When we replace the rule \ref{eq12:r23} by
\begin{alignat}{1}
  \label{eq12:r23p}\tag{$r_{23}'$}
  & p \leftarrow \asum\lits{2 : x_1, 2 : x_2, 3 : x_3} \neq 6
\end{alignat}
to obtain $\prg_5'$,
there is no interpretation $Z\subseteq X$ such that
$\eval{\asum\lits{2 : x_1, 2 : x_2, 3 : x_3} = 6}{Z} =\nolinebreak 6$.
In turn, we have that
$Z\models \asum\lits{2 : x_1,\linebreak[1] 2 : x_2,\linebreak[1] 3 : x_3} \neq 6$
for every $Z\subseteq X$,
which means that the condition of
$(\emptyset,X)\sat{\lpst} \body{$\ref{eq12:r23p}$}$ applies and
$\tpu{\lpst}{\prg_5'}{X}{1}=\tpc{\lpst}{\prg_5'}{X}{\emptyset}=\{p\}$.
Given that $\eval{\asum\lits{1:p} > 0}{\linebreak[1]Z}=1$
for all interpretations $\{p\}\subseteq Z$,
we get $(\{p\},X)\sat{\lpst} \body{r}$ for every $r\in\prg_5'$,
which leads to
$\tpu{\lpst}{\prg_5'}{X}{\infty}=
 \tpu{\lpst}{\prg_5'}{X}{2}=
 \tpc{\lpst}{\prg_5'}{X}{\{p\}}=
 \{x_1,x_2,x_3,p\}= X$.
This shows that $X$ is an \lpst-answer set of $\prg_5'$.

The programs $\prg_5$ and $\prg_5'$ illustrate that the \checkas task
for \lpst-answer sets can be used to decide whether an
instance of the \NP-complete \emph{subset sum} problem
is (un)satisfiable,
as programs following the same scheme along with an interpretation
consisting of all atoms capture the complement of \emph{subset sum}
for arbitrary multisets of weights and bounds.
Moreover, Section~\ref{subsec:satisfiability} discusses reductions
of \emph{subset sum} and the likewise \NP-complete \emph{subset product} problem
to the satisfiability of aggregate expressions with \aavg or \atimes, respectively,
so that their complementary aggregate expressions with $\neq$ as comparison operator
also make the \checkas task \coNP-complete.%
\end{example}

For a program \prg with arbitrary aggregate expressions and a given model $X$ of \prg,
the \checkas task of deciding whether $X$ is an \mr-answer set of \prg
can be accomplished by \begin{rev}guessing\end{rev}, for each aggregate expression
$A\in\bigcup_{r\in\reduct{\prg}{X}}\body{r}$,
an interpretation $Z_A\subseteq X$ such that $Z_A\models A$.
Given that $X\models A$, such an interpretation $Z_A$ exists for
every $A$ under consideration and can be used to witness that
$(Y,X)\sat{\mr} A$ for interpretations $Z_A\subseteq Y\subseteq X$.
Checking whether $Z_A\subseteq Y$ instead of taking the condition of $(Y,X)\sat{\mr} A$
as such, in order to approximate
$\tpc{\mr}{\prg}{X}{Y}$ for some $Y\subseteq X$,
yields a tractable immediate consequence operator,
whose least fixpoint matches $\tpu{\mr}{\prg}{X}{\infty}$ for well-chosen
interpretations $Z_A$.
This shows that the \checkas task for \mr-answer sets stays in \NP
also in the presence of non-convex aggregate expressions.

For establishing \NP-hardness, 
the complexity of checking the existence of some $Z\subset\atoms{A}\cap X$ such that
$Z\models A$ for an aggregate expression $A$ with $X\models A$ matters, as the
construction of $\tpu{\mr}{\prg}{X}{\infty}$ gets tractable and
the complexity of \checkas drops to \poly when the satisfiability of
aggregate expressions (by some smaller interpretation) can be
determined in polynomial time.
As the discussion in Section~\ref{subsec:satisfiability} shows,
deciding satisfiability is \NP-hard only for aggregate expressions with
\asum, \atimes, or \aavg along with
the comparison operator~$=$,
so that any other (non-convex) aggregate expressions do not make the \checkas task
\NP-complete.
However, for an aggregate expression $A$ of the form
$\atimes\lits{w_1:p_1,\linebreak[1]\dots,\linebreak[1]w_n:p_n}=\bound$ such that
$\bound=0$, only singletons $Z=\{p\}$ with $\eval{A}{Z}=0$
are relevant as subsets $Z\subset\atoms{A}\cap X$ such that $Z\models A$.
Otherwise, if $w_0\neq 0$, 
we have to guarantee that
$\eval{A}{X\setminus Z}=1$,
so that a linear collection of ($\subseteq$-minimal) subsets $Z\subset \atoms{A}\cap X$
with $Z\models A$ is obtained
by excluding atoms $p\in X$ such that $\eval{A}{\{p\}}=1$ together with
a maximum even number of atoms $p\in X$ such that $\eval{A}{\{p\}}=-1$.
As there are at most $|\atoms{A}\cap X|$ relevant $Z\subset \atoms{A}\cap X$ 
in either case, aggregate expressions with \atimes do not yield \NP-completeness
of the \checkas task for \mr-answer sets.
The next example illustrates that this is different for
aggregate expressions with \asum or \aavg
along with the comparison operator~$=$.

\begin{example}\label{ex:checkmr}
Let us investigate the model $X=\{x_1,x_2,x_3,p\}$ of the following program~$\prg_6$:\pagebreak[1]
\begin{alignat}{1}
  \label{eq13:r24}\tag{$r_{24}$}
  & x_1 \leftarrow \top
  \\
  \label{eq13:r25}\tag{$r_{25}$}
  & x_2 \leftarrow \top
  \\
  \label{eq13:r26}\tag{$r_{26}$}
  & x_3 \leftarrow \top
  \\
  \label{eq13:r27}\tag{$r_{27}$}
  & p \leftarrow \asum\lits{2 : x_1, 2 : x_2, 3 : x_3, -2 : p} = 5
\end{alignat}
In order to decide whether $X$ is an \mr-answer set of $\prg_6$, we need to
construct $\tpu{\mr}{\prg_6}{X}{\infty}$,
where
$\tpu{\mr}{\prg_6}{X}{1}=\tpc{\mr}{\prg_6}{X}{\emptyset}=\{x_1,x_2,x_3\}$
in view of the facts
\ref{eq13:r24}, \ref{eq13:r25}, and \ref{eq13:r26}.
Then, the condition of $(\{x_1,x_2,x_3\},X)\sat{\mr} \body{$\ref{eq13:r27}$}$
is fulfilled if there is some interpretation
$Z\subseteq \{x_1,x_2,x_3\}$ such that
$\eval{\asum\lits{2 : x_1,\linebreak[1] 2 : x_2,\linebreak[1] 3 : x_3,\linebreak[1] -2 : p} = 5}{Z} = 5$.
This is the case for $Z=\{x_1,x_3\}$ (and $Z=\{x_2,x_3\}$), so that
\begin{alignat*}{1}
  \tpu{\mr}{\prg_6}{X}{\infty}=\tpu{\mr}{\prg_6}{X}{2}=\tpc{\mr}{\prg_6}{X}{\{x_1,x_2,x_3\}}=
 \{x_1,x_2,x_3,p\}=X\text{.}
\end{alignat*}

When we replace the rule \ref{eq13:r27} by
\begin{alignat}{1}
  \label{eq13:r27p}\tag{$r_{27}'$}
  & p \leftarrow \asum\lits{2 : x_1, 2 : x_2, 3 : x_3, -1 : p} = 6
\end{alignat}
in the modified program $\prg_6'$,
there is no $Z\subseteq \{x_1,x_2,x_3\}$ such that
$\eval{\asum\lits{2 : x_1,\linebreak[1] 2 : x_2,\linebreak[1] 3 : x_3,\linebreak[1] -1 : p} = 6}{Z} = 6$,
so that
$\tpu{\mr}{\prg_6'}{X}{\infty}=\tpu{\mr}{\prg_6'}{X}{1}=\{x_1,x_2,x_3\}$
disproves $X$ to be an \mr-answer set of $\prg_6'$.
Both programs $\prg_6$ and $\prg_6'$
are devised such that the construction of
$\tpu{\mr}{\prg_6}{X}{\infty}$ or $\tpu{\mr}{\prg_6'}{X}{\infty}$, respectively,
involves checking whether an instance of the \emph{subset sum} problem is satisfiable.
If so,
the weight associated with the atom $p$ to be concluded
is set such that the desired bound is obtained by summing up all weights.
As programs following the same scheme along with an interpretation
consisting of all atoms can be used to decide arbitrary instances
of \emph{subset sum},
the \checkas task for \mr-answer sets is \NP-complete in the
presence of non-convex aggregate expressions of the form
$\asum\lits{w_1:p_1,\linebreak[1]\dots,\linebreak[1]w_n:p_n}=\bound$.

Rules like \ref{eq13:r27} and \ref{eq13:r27p} can also be adjusted
to use the aggregation function \aavg instead of \asum by introducing
an additional fact $x\leftarrow \top$ and taking
the inverse of the original bound as weight for~$x$:\pagebreak[1]%
\begin{alignat}{1}
  \label{eq13:r28}\tag{$r_{28}$}
  & p \leftarrow \aavg\lits{2 : x_1, 2 : x_2, 3 : x_3, -5 : x, -2 : p} = 0
  \\
  \label{eq13:r28p}\tag{$r_{28}'$}
  & p \leftarrow \aavg\lits{2 : x_1, 2 : x_2, 3 : x_3, -6 : x, -1 : p} = 0
\end{alignat}
The condition of
$(\{x_1,x_2,x_3,x\},X\cup\{x\})\sat{\mr} \body{$\ref{eq13:r28}$}$ or
$(\{x_1,x_2,x_3,x\},X\cup\{x\})\sat{\mr} \body{$\ref{eq13:r28p}$}$, respectively,
then again amounts to deciding the satisfiability of an underlying instance of
\emph{subset sum}.
Hence, we have that the \checkas\ task for \mr-answer sets is likewise \NP-complete for
programs including non-convex aggregate expressions of the form
$\aavg\lits{w_1:p_1,\linebreak[1]\dots,\linebreak[1]w_n:p_n}=\bound$.%
\end{example}

\subsection{Answer Set Existence}\label{subsec:exist}

The \existas reasoning task addresses
the decision problem of whether a program \prg has some
\any-answer set for $\any\in\{\fflp,\linebreak[1]\gz,\linebreak[1]\lpst,\linebreak[1]\mr,\linebreak[1]\pdb\}$,
i.e., a decision problem similar to \checkas but having no fixed interpretation $X\subseteq \at$.
The summary of completeness properties for particular complexity classes
is provided in Table~\ref{tab:complexity},
where completeness now also applies for \poly in view of the well-known \poly-completeness of
deciding whether the $\subseteq$-minimal model of a positive program
includes some atom of interest \begin{rev}\cite{daeigovo01a}\end{rev}.
In fact, when using (monotone) aggregate expressions of the form $\asum\lits{1:p}>0$
to represent propositional atoms~$p$ in the bodies of rules, all \any-answer set semantics
reproduce the $\subseteq$-minimal model of a positive program as the unique \any-answer set.
Given that our notion of non-disjunctive programs syntactically allows for constraints,
a program does not necessarily have a
($\subseteq$-minimal) model though, even when all aggregate expressions are monotone,
which makes the \existas task non-trivial and thus \poly-complete.
While going beyond the simple form $\asum\lits{1:p}>0$ of monotone aggregate expressions
lets the \gz-answer set semantics diverge from the other aggregate semantics,
as observed on Example~\ref{ex:gz},
such additional syntactic freedom does not increase the computational complexity
any further because the \any-answer set semantics,
for $\any\in\{\fflp,\linebreak[1]\lpst,\linebreak[1]\mr,\linebreak[1]\pdb\}$,
still coincide and the constructions by means of different immediate consequence operators
remain tractable.

Regarding anti-monotone and convex aggregate expressions,
the rows for the \existas task in Table~\ref{tab:complexity}
follow the pattern of \checkas,
where \poly-membership or \coNP-completeness, respectively, of \checkas leads to
\NP- or \stwo-completeness of \existas.
For \NP-hardness,
anti-monotone aggregate expressions of the form $\asum\lits{1:p}<1$ are sufficient,
as they allow for representing negated propositional atoms~$p$ in rule bodies,
which directly lead to elevated complexity of deciding whether some
(standard) answer set exists \cite{martru91c}.
Likewise, the \stwo-hardness of \existas for \pdb-answer sets
follows from the simultaneous availability of
monotone and anti-monotone
aggregate expressions of the form $\asum\lits{1:p}>0$ or $\asum\lits{1:p}<1$,
as the reduction from quantified Boolean formulas (with one quantifier alternation)
by \citeN{dematr04a} shows.

The additional availability of non-convex aggregate expressions
in programs with arbitrary aggregate expressions
does not increase the complexity of \existas beyond \NP for
\gz- as well as \mr-answer sets.
Regarding \gz-answer sets, 
the strong condition $\atoms{A}\cap Y = \atoms{A}\cap X$ for $(Y,X)\sat{\gz} A$
is unaffected by the monotonicity of an aggregate expression~$A$ \cite{alvleo15a},
as already observed on the tractability of the \checkas task.
The latter reasoning task happens to be in \NP for \mr-answer sets,
so that deciding whether a program including non-convex aggregate expressions
has some \mr-answer set does not involve orthogonal combinatorial problems.
Unlike that, the \stwo-hardness of \existas for \pdb-answer sets is
established already for programs with convex aggregate expressions,
and non-convex aggregate expressions do not make a difference here
because neither determining candidate models $X$ of a program \prg
nor checking whether $\tpu{\pdb}{\prg}{X}{\infty}=X$ becomes more complex
in their presence.

The complexity of \existas for \fflp-answer sets correlates to the monotonicity
of aggregate expressions,
given the same consideration as for \checkas that (proper) disjunctive rules can be
captured in terms of non-disjunctive programs including non-convex aggregate expressions
\cite{ferraris11a}.
This yields \stwo-completeness of the \existas task
for \fflp-answer sets of programs with non-convex aggregate expressions
regardless of their particular 
aggregation functions, comparison operators,
weights and bounds \cite{alvfab13a}.
While the complexity of \existas is in general the same for \lpst-answer sets,
the \stwo-hardness again depends on the complexity of deciding the satisfiability
of the complement of a (non-convex) aggregate expression~$A$.
That is, the complexity drops to \NP without aggregate expressions
$\agg\lits{w_1:p_1,\linebreak[1]\dots,\linebreak[1]w_n:p_n}\neq\bound$
such that $\agg\in\{\asum,\linebreak[1]\atimes,\linebreak[1]\aavg\}$
because the underlying \checkas task for \lpst-answer sets gets tractable when
programs do not involve aggregate expressions of this form.
As further detailed in the next example,
aggregate expressions with \asum or \aavg along with the comparison operator~$\neq$
allow for expressing the \stwo-complete \emph{generalized subset sum} problem
\cite{bekalaplry02a}, so that the same complexity is obtained for
\lpst-answer sets of programs including such non-convex aggregate expressions.
Although we are unaware of literature addressing the complexity of a
corresponding generalized version of the \emph{subset product} problem,
the \coNP-hardness of the \checkas\ task
suggests that \existas remains \stwo-hard with
\atimes as well.

\begin{example}\label{ex:exist}
As checked in Example~\ref{ex:subsum} and Example~\ref{ex:mr},
the interpretation $X=\{y_1,x_2,z_1,z_2,p\}$
is the unique \any-answer set, for $\any\in\{\fflp,\lpst,\pdb\}$, of
the program~$\prg_3$:\pagebreak[1]%
\begin{alignat}{1}
  \tag{\ref{eq6:r6}}
  & x_1 \leftarrow \asum\lits{1:y_1} < 1
  \\
  \tag{\ref{eq6:r7}}
  & y_1 \leftarrow \asum\lits{1:x_1} < 1
  \\
  \tag{\ref{eq6:r8}}
  & x_2 \leftarrow \asum\lits{1:y_2} < 1
  \\
  \tag{\ref{eq6:r9}}
  & y_2 \leftarrow \asum\lits{1:x_2} < 1
  \\
  \tag{\ref{eq6:r10}}
  & z_1 \leftarrow \asum\lits{1:p} > 0
  \\
  \tag{\ref{eq6:r11}}
  & z_2 \leftarrow \asum\lits{1:p} > 0
  \\
  \tag{\ref{eq6:r12}}
  & p \leftarrow \asum\lits{1:y_1,2:y_2,2:z_1,3:z_2} \neq 5
  \\
  \tag{\ref{eq6:r13}}
  & \bot \leftarrow \asum\lits{1:p} < 1
\end{alignat}
In fact, $\{y_1,y_2\}\cap X=\{y_1\}$ represents the single solution
to the instance
\begin{alignat*}{1}
  \exists y_1 y_2 \forall z_1 z_2
 (1\cdot y_1+\linebreak[1]
  2\cdot y_2+\linebreak[1]
  2\cdot z_1+\linebreak[1]
  3\cdot z_2 \neq 5)
\end{alignat*}
of the \emph{generalized subset sum} problem.
The rules \ref{eq6:r6}, \ref{eq6:r7}, \ref{eq6:r8}, and \ref{eq6:r9}
make use of the auxiliary atoms $x_1$ and $x_2$ to express 
exclusive choices between $x_1$ and $y_1$ as well as $x_2$ and~$y_2$,
where the chosen atoms $y_i$ stand for a solution candidate.
If the candidate is indeed a solution, any truth assignment of the atoms $z_i$
leads to a weighted sum that is different from the bound,
which is $5$ for our instance.
In this case,
we have that $Z\models \body{$\ref{eq6:r12}$}$
for every interpretation $\{y_1,y_2\}\cap X\subseteq Z \subseteq X$,
so that the head atom $p$ must necessarily be true and is also concluded by
the operators $\tpx{\lpst}{\prg_3}{X}$ as well as $\tpx{\pdb}{\prg_3}{X}$.
The remaining rules \ref{eq6:r10} and \ref{eq6:r11} establish that the
atoms $z_i$ follow from $p$, which makes sure that a potential counterexample, i.e.,
some truth assignment of the atoms $z_i$ such that the weighted sum matches the given bound,
corresponds to a smaller interpretation $Y\subset X$ disproving $X$ to be a \any-answer set of~$\prg_3$.
Finally, the constraint \ref{eq6:r13} expresses that
$p$ must be true, while its provability relies on~\ref{eq6:r12},
and thus eliminates any solution candidate that would directly give the bound as weighted sum
when all of the atoms $z_i$ are assigned to false.

Introducing
an additional fact $x\leftarrow \top$ and replacing \ref{eq6:r12} by%
\begin{alignat}{1}
  \label{eq6:r12p}\tag{$r_{12}'$}
  & p \leftarrow \aavg\lits{1:y_1,2:y_2,2:z_1,3:z_2,-5:x} \neq 0
\end{alignat}
leads to a modified program $\prg_3'$ including a non-convex aggregate expression with
\aavg instead of \asum.
This aggregate expression takes the inverse of the original bound as weight for a
necessarily true atom~$x$,
so that the average happens to be $0$ \ifaoif
the weighted sum over the remaining (true) atoms matches the bound.
Hence, we have that \any-answer sets $X\cup\{x\}$ of $\prg_3'$ correspond to
\any-answer sets $X$ of $\prg_3$ for $\any\in\{\fflp,\lpst,\pdb\}$.

Given that programs following the scheme of $\prg_3$ or~$\prg_3'$, respectively,
allow for expressing arbitrary instances of \emph{generalized subset sum},
we conclude that the \existas task for the three aggregate semantics of interest, in particular,
\lpst-answer sets of programs with non-convex aggregate expressions as used in $\prg_3$
and~$\prg_3'$, is \stwo-hard.
Such elevated complexity is not obtained for \gz- and \mr-answer sets,
and as discussed in Example~\ref{ex:mr},
the program $\prg_3$ does not have any \gz-answer set although the corresponding instance of
\emph{generalized subset sum} is satisfiable,
while $X$ and two more interpretations that do not represent solutions
are \mr-answer sets of~$\prg_3$.
\end{example}

Finally, note that
the computational complexity of the \checkas and \existas reasoning tasks
is a common subject of investigation for specific logic programs.
In this regard, the comparably low complexity for \gz-answer sets does not
come as a surprise, as avoiding so-called vicious circles \cite{gelzha19a}
circumvents elevated complexity due to (sophisticated) aggregate expressions.
Hardness properties concerning \pdb-answer sets are immediate consequences of
the simultaneous availability of monotone and anti-monotone aggregate expressions,
given the computational complexity of the so-called ultimate semantics \cite{dematr04a},
and the complexity of reasoning tasks for \fflp-answer sets has been thoroughly
studied in the corresponding literature \cite{alvfab13a,fapfle11a,ferraris11a}.
For \lpst-answer sets,
the elevated complexity due to aggregate expressions of the form
$\asum\lits{w_1:p_1,\linebreak[1]\dots,\linebreak[1]w_n:p_n}\neq\bound$ or
$\aavg\lits{w_1:p_1,\linebreak[1]\dots,\linebreak[1]w_n:p_n}\neq\bound$
has been put forward by \citeN{sonpon07b}, and we here add
$\atimes\lits{w_1:p_1,\linebreak[1]\dots,\linebreak[1]w_n:p_n}\neq\bound$
to these considerations.
To our knowledge,
the computational complexity of reasoning tasks for \mr-answer sets has 
not been investigated in depth, and the \NP-hardness of \checkas
for programs with aggregate expressions 
$\asum\lits{w_1:p_1,\linebreak[1]\dots,\linebreak[1]w_n:p_n}=\bound$ or
$\aavg\lits{w_1:p_1,\linebreak[1]\dots,\linebreak[1]w_n:p_n}=\bound$
was unrecognized before.


\section{Discussion}\label{sec:discussion}

As the variety of proposals investigated in Section~\ref{sec:answersets} shows,
the design of a general semantics of aggregates and efficient implementations
thereof have been long-standing challenges.
The intricacy results from two particularities:
unlike for conjunctions of literals,
aggregate expressions can yield a non-convex satisfiability pattern,
and the handling of (explicit) negation by the \nap connective requires additional care.
In the following, we discuss such phenomena,
related rewriting methods and first-order semantics
used to implement or generalize aggregates, respectively,
and further extensions of logic programs to custom aggregate expressions.

The pioneering \textsc{smodels} system \cite{siniso02a} handles 
\emph{weight constraints} 
$\asum\lits{w_1:\nolinebreak \lit_1,\linebreak[1]\dots,\linebreak[1]w_n:\lit_n}\geq\bound$
such that each $\lit_i$, for $1\leq i\leq n$,
is a literal of the form
$\lit_i=p$ or $\lit_i=\naf{p}$ over some atom $p\in\at$.
While negative weights $w_i$ can be supplied in the input,
weighted literals $w_i:\lit_i$ 
are transformed according to\pagebreak[1]
\begin{alignat*}{1}
\nwt{w_i:\lit_i} = \left\{
           \begin{array}{@{}l@{\,}l@{}}
           w_i:\lit_i
             & \text{, if $w_i \geq 0$,}
           \\
          -w_i:\opp{\lit_i}
             & \text{, if $w_i < 0$,}
           \end{array}
           \right.
\end{alignat*}
where $\opp{\lit_i}=\naf{p}$, if $\lit_i=p$,
or $\opp{\lit_i}=p$, if $\lit_i=\naf{p}$, denotes the complement of a literal~$\lit_i$,
in the mapping to
$\asum[\nwt{w_i:\lit_i} \mid 1\leq i\leq n]\geq\bound-\sum_{1\leq i\leq n,w_i<0}w_i$,
based on the idea that sanctioning some literal by a negative weight is in terms of
satisfiability the same as associating its complement with the corresponding positive amount.
Given that the reduct adopted by \textsc{smodels} evaluates negative literals,
the transformed weight constraints 
without negative weights become
monotone in the context of checking the provability of true atoms.
Hence, deciding whether a given interpretation is an answer set
gets tractable, which diverges from the \coNP-completeness of
\fflp-answer set checking for programs with non-convex aggregate expressions
(cf.\ Table~\ref{tab:complexity}),
as already established in the absence of negative literals built by means of the \nap connective.
This complexity gap brings about peculiarities, considering that $\{p\}$ is 
an answer set produced by \textsc{smodels} as well as a
\any-answer set, for $\any\in\{\fflp,\linebreak[1]\lpst,\linebreak[1]\mr,\linebreak[1]\pdb\}$,
of a program with just the rule $p\leftarrow \asum\lits{0:p}\geq 0$.
When the rule is changed to $p\leftarrow \asum\lits{1:p,-1:p}\geq 0$,
we have that $\{p\}$ remains a \any-answer set,
while \textsc{smodels}
transforms the rule to
$p\leftarrow \asum\lits{1:p,1:\naf{p}}\geq 1$
and takes
$p\leftarrow \asum\lits{1:p}\geq 1$
as reduct relative to the unique model $\{p\}$ \begin{rev}---
In practice, negative weights are eliminated by the front-end
\textsc{lparse} \cite{lparseManual} during grounding, which then
passes the transformed weight constraints 
on to \textsc{smodels} or other ASP solvers\end{rev}.
This yields the empty interpretation~$\emptyset$ as a smaller model
disproving $\{p\}$,
so that the transformation by \textsc{smodels} leads to different semantics of the rules
$p\leftarrow \asum\lits{0:p}\geq 0$ and
$p\leftarrow \asum\lits{1:p,-1:p}\geq 0$.
A common issue about transformations of logic programs, in general, and of aggregate expressions,
in particular, is that satisfiability-preservation does not necessarily yield equitable
provability of atoms.
For guaranteeing replaceability regardless of the specific program context,
strong equivalence \cite{lipeva01a}
needs to be established, and \begin{rev}\citeN{bojani20a}, \citeN{ferraris11a}, and \citeN{gehakalisc15a}\end{rev}
investigate this concept for aggregate expressions.

According to \cite{ferlif02a},
a weight constraint 
$\asum\lits{w_1:\lit_1,\dots,w_n:\lit_n}\geq\bound$ with positive weights $w_i$, for $1\leq i\leq n$,
is equivalent to 
\begin{alignat*}{1}
\mbox{$\bigvee$}_{I\subseteq\{1,\dots,n\},\sum_{i\in I}w_i\geq\bound}\mbox{$\bigwedge$}_{i\in I}\lit_i\text{,}
\end{alignat*}
where the disjunction can be understood as a shorthand for several alternative rule bodies.
Similar unfoldings of aggregate expressions have been 
investigated \begin{rev}by \citeN{pedebr03a} and \citeN{sopoel06a},\end{rev} and yield
semantic correspondences to \lpst-answer sets \cite{sonpon07b}.
More compact rewritings of weight constraints to aggregate-free rules,
devised for increasing the range of applicable solving systems,
are based on sequential weight counters \cite{ferlif02a} or merge-sorting \cite{bogeja14a},
inspired by corresponding encodings of pseudo-Boolean constraints
in propositional logic \cite{baboro09a,homast12a,rouman09a}.
In general, when the joint use of positive and negative weights makes a weight constraint non-convex,
the more sophisticated formula
\begin{alignat*}{1}
 \mbox{$\bigwedge$}_{I\subseteq\{1,\dots,n\},\sum_{i\in I}w_i<\bound}
 \left(\mbox{$\bigwedge$}_{i\in I}\lit_i
  \rightarrow
  \mbox{$\bigvee$}_{i\in \{1,\dots,n\}\setminus I}\lit_i
 \right)
\end{alignat*}
has been shown to be semantic-preserving \cite{ferraris11a}.
This scheme introduces nested implications, which
lead to elevated complexity of reasoning tasks, as discussed in Section~\ref{subsec:check}.
Such complex constructs are not directly supported by the ASP systems 
\textsc{clingo} and \textsc{dlv},
whose propagation procedures \cite{fapfledeie08a,gekakasc09a}
are designed for (anti\nobreakdash-)monotone aggregate expressions ---
A convex aggregate expression can be
decomposed into the conjunction
of a monotone and an anti-monotone aggregate expression
\cite{liutru06a}, e.g.,
by syntactically taking the comparison operators $\leq$ and $\geq$
to represent~$=$.
Reasoning on non-convex aggregate expressions can still be accomplished by means of a
rewriting to disjunctive rules with monotone aggregate expressions \cite{alfage15a},
where (proper) disjunctive rules are needed only if a non-convex aggregate expression
occurs in positively recursive rules, which are necessary to express a \stwo-hard
problem like \emph{generalized subset sum} in Example~\ref{ex:exist}.
Since this rewriting is implemented in the grounding procedure of \textsc{clingo}
\cite{gekasc15a},
any disjunctive ASP solver that handles recursive monotone aggregate
expressions, e.g., used to encode \emph{company controls},
can be applied to reason on non-convex aggregate expressions as well.
\begin{rev}
The input language of \textsc{clingo} \cite{gehakalisc15a}, however, excludes the
aggregation functions \atimes and \aavg,
which are also not part of the \aspcore specification
of a common first-order language for ASP systems \cite{cafageiakakrlemarisc19a}\end{rev}.
Moreover, \textsc{clingo} rewrites aggregate expressions with \amin or \amax to aggregate-free rules,
so that only \acount and \asum effectively lead to
monotone aggregate expressions that use the \asum aggregation function at the ground level.

Current ASP solvers deal with weight constraints
$\asum\lits{w_1:\lit_1,\dots,w_n:\lit_n}\geq\bound$ with positive weights~$w_i$,
and grounding (components of) systems like \textsc{clingo} \cite{gekasc15a},
\textsc{dlv} \cite{cadofupeza20a}, and \textsc{lparse} \cite{lparseManual}
establish such a format, using transformations like those given \begin{rev}by \citeN{alfage15a}\end{rev}.
In the context of non-disjunctive programs,
translational approaches to propositional logic \cite{jannie11a} rely on
the aforementioned rewritings to aggregate-free rules \cite{ferlif02a,bogeja14a},
and even more compact, linear representations in terms of pseudo-Boolean constraints
or integer programming \cite{gejari14c} enable the use of corresponding back-end solvers
to compute answer sets.
The ASP systems \textsc{clingo}, \textsc{dlv}, and \textsc{smodels}
incorporate propagation procedures 
\cite{fapfledeie08a,gekakasc09a,siniso02a}
that extend Boolean constraint propagation \cite{rouman09a}
from pseudo-Boolean constraints to logic programs.
Their restrictions are that
\textsc{smodels} handles non-disjunctive programs only,
aggregate expressions must be non-recursive for \textsc{dlv},
and \textsc{clingo} does not use compact data structures for aggregate expressions
in positively recursive rules such that their head atoms occur together in some
(proper) disjunctive rule head
(i.e., the program part under consideration is not head-cycle-free \cite{bendec94a}),
while more space-consuming rewriting to aggregate-free rules is performed otherwise.
Moreover,
the recent ASP solver \textsc{wasp} \cite{alamdolemari19a} implements a collective
propagation procedure \cite{aldoma18a} for aggregate expressions
$\asum\lits{w_1:\lit_1,\dots,w_n:\lit_n}\geq\bound$
with the same weighted literals $w_1:\lit_1,\dots,w_n:\lit_n$ and
different bounds \bound,
and the lazy-grounding ASP system \textsc{alpha} \cite{wetafr20a}
features an incremental on-demand rewriting \cite{bojawe19a} to aggregate-free rules.

While the semantics by \citeN{fapfle11a} and by \citeN{ferraris11a} agree
for the syntax of aggregate expressions considered in this survey, i.e.,
aggregates over propositional atoms,
the different reduct notions, which either eliminate falsified expressions and thus
negative literals or not, lead to distinct outcomes in the presence of \nap.
For instance, the reduct of 
$p\leftarrow \asum\lits{1: p,\linebreak[1]1:\naf{p}}\geq 1$
relative to the interpretation $\{p\}$ is
the rule itself according to \begin{rev}\citeN{fapfle11a}\end{rev},
and $p\leftarrow \asum\lits{1:p}\geq 1$ by \begin{rev}Ferraris' \citeyear{ferraris11a} definition\end{rev}. 
Hence, either $\{p\}$ or $\emptyset$ is obtained as $\subseteq$-minimal model of the reduct,
so that the semantics disagree about whether $\{p\}$ is an answer set.
Similarly, $p\leftarrow \naf{\asum\lits{1: p}< 1}$ yields
the rule as such or $p\leftarrow \top$ as the reduct relative to $\{p\}$,
resulting in either $\emptyset$ or $\{p\}$ as $\subseteq$-minimal model.
That is, answer sets according to \begin{rev}\citeN{fapfle11a} and \citeN{ferraris11a}\end{rev}
can be mutually distinct when the \nap connective is used in front of
aggregate expressions or propositional atoms subject to aggregation functions.
To circumvent such discrepancies, the
\aspcore \begin{rev}language specification\end{rev} \cite{cafageiakakrlemarisc19a} requires 
aggregate-stratification \cite{fapfledeie08a},
which restricts occurrences of
aggregate expressions in a program to be non-recursive.
Under this condition,
apart from \pdb-answer sets that differ already for aggregate-free programs \cite{depebr01a,dematr04a},
the aggregate semantics investigated in Section~\ref{sec:answersets} agree,
and aggregate expressions may not increase the computational complexity of reasoning tasks.
\citeN{harlif19a} relax the aggregate-stratification condition and show that
answer sets according to \begin{rev}\citeN{fapfle11a} and \citeN{ferraris11a}\end{rev} coincide
when recursive aggregate expressions do not involve the \nap connective, i.e.,
they are of the form~\eqref{eq:syntax:agg},
while no syntactic restrictions are imposed otherwise, e.g.,
for aggregate expressions occurring in constraints.
Notably, the correspondence in \cite{harlif19a} is established at the first-order level,
thus connecting prior first-order generalizations of the two semantics \cite{baleme11a,gehakalisc15a}
based on second-order logic or infinitary formulas, respectively.
Moreover, answer sets according to \begin{rev}\citeN{fapfle11a} and \citeN{ferraris11a}\end{rev}
can be characterized in terms of each other \cite{leemen09a,truszczynski10a},
so that their expressiveness is the same regardless of
semantic differences arising on the \nap connective.

Beyond their use for a compact representation of properties on sets of atoms,
semantics for aggregate expressions have been taken as basis for defining
the answer sets of logic programs with further extensions,
such as description logic and higher-order logic programs with external atoms
\cite{shwaeifirekrde14a}.
Similar to aggregate expressions,
monotonicity properties of external atoms affect reasoning about them,
and dedicated solving techniques take advantage of so-called assignment-monotonicity
\cite{eikarewe18a}.
Extensions of ASP systems by theory propagators
\cite{cudorisc20a,jakaosscscwa17a,liesus17a}
likewise interpret specific atoms as custom aggregate expressions
and incorporate respective procedures for propagating their truth values.
In this broad sense, the notion of an aggregate includes any method of evaluating
atoms as a compound, where some frequently used aggregation functions, in particular,
the \asum aggregation function, are accommodated off-the-shelf in the modeling language of ASP.

While there is already a considerable body of work on aggregates, there clearly are numerous open issues to be addressed.
We would like to outline some of them, of course without any claim of completeness.
First of all, identifying sublanguages on which different semantics coincide is still a relevant topic. As mentioned earlier, most semantics agree on aggregate-stratified programs, and many also correspond for programs with convex aggregates.
Another issue is that encodings with unstratified aggregate occurrences are fairly rare at the moment, especially when considering ``real-world''  applications. We believe, however, that providing a meaningful semantics for language constructs is important also when there are no frequent applications for them. Nevertheless, potential application fields might lie, for instance, in the area of analysis of dynamic systems, e.g., addressing the location of fixpoints in biological systems.
A related question is whether the proposed languages are actually fit for various practical purposes. We believe that postulating formal properties that semantics for programs with aggregates should satisfy is important and still lacking. For example, formalizations of the Closed World Assumption \begin{rev}\cite{reiter77a}\end{rev} for programs with aggregates would be of interest. There is also work arguing that certain programs are not handled well by the existing semantics, yet suggestions as, e.g., \begin{rev}by \citeN{alvfab19}\end{rev} for ``repairing'' existing approaches to cover such cases do not seem satisfactory so far. Finally, we would like to mention that more implementations are needed. \begin{rev}Most\end{rev} proposed semantics have never been supported by any system, making comparisons difficult and applications impossible\begin{rev}, and merely \textsc{clingo} implements non-convex aggregate expressions in positively recursive rules (under \fflp-answer set semantics). Hence, even\end{rev} implementations that are not at all geared towards efficiency would be useful. In summary, there is a substantial amount of room for future work on aggregates in Answer Set Programming.%



\begin{thebibliography}{}

\bibitem[\protect\citeauthoryear{Alviano, Amendola, Dodaro, Leone, Maratea, and
  Ricca}{Alviano et~al\mbox{.}}{2019}]{alamdolemari19a}
{\sc Alviano, M.}, {\sc Amendola, G.}, {\sc Dodaro, C.}, {\sc Leone, N.}, {\sc
  Maratea, M.}, {\sc and} {\sc Ricca, F.} 2019.
\newblock Evaluation of disjunctive programs in {WASP}.
\newblock In {\em Proceedings of the Fifteenth International Conference on
  Logic Programming and Nonmonotonic Reasoning (LPNMR'19)}, {M.~Balduccini},
  {Y.~Lierler}, {and} {S.~Woltran}, Eds. Lecture Notes in Artificial
  Intelligence, vol. 11481. Springer-Verlag, 241--255.

\bibitem[\protect\citeauthoryear{Alviano, Calimeri, Dodaro, Fusc{\`a}, Leone,
  Perri, Ricca, Veltri, and Zangari}{Alviano
  et~al\mbox{.}}{2017}]{alcadofuleperiveza17a}
{\sc Alviano, M.}, {\sc Calimeri, F.}, {\sc Dodaro, C.}, {\sc Fusc{\`a}, D.},
  {\sc Leone, N.}, {\sc Perri, S.}, {\sc Ricca, F.}, {\sc Veltri, P.}, {\sc
  and} {\sc Zangari, J.} 2017.
\newblock The {ASP} system {DLV}2.
\newblock In {\em Proceedings of the Fourteenth International Conference on
  Logic Programming and Nonmonotonic Reasoning (LPNMR'17)}, {M.~Balduccini}
  {and} {T.~Janhunen}, Eds. Lecture Notes in Artificial Intelligence, vol.
  10377. Springer-Verlag, 215--221.

\bibitem[\protect\citeauthoryear{Alviano, Dodaro, and Maratea}{Alviano
  et~al\mbox{.}}{2018}]{aldoma18a}
{\sc Alviano, M.}, {\sc Dodaro, C.}, {\sc and} {\sc Maratea, M.} 2018.
\newblock Shared aggregate sets in answer set programming.
\newblock {\em Theory and Practice of Logic Programming\/}~{\em 18,\/}~3-4,
  301--318.

\bibitem[\protect\citeauthoryear{Alviano and Faber}{Alviano and
  Faber}{2013}]{alvfab13a}
{\sc Alviano, M.} {\sc and} {\sc Faber, W.} 2013.
\newblock The complexity boundary of answer set programming with generalized
  atoms under the {FLP} semantics.
\newblock In {\em Proceedings of the Twelfth International Conference on Logic
  Programming and Nonmonotonic Reasoning (LPNMR'13)}, {P.~Cabalar} {and}
  {T.~Son}, Eds. Lecture Notes in Artificial Intelligence, vol. 8148.
  Springer-Verlag, 67--72.

\bibitem[\protect\citeauthoryear{Alviano and Faber}{Alviano and
  Faber}{2019}]{alvfab19}
{\sc Alviano, M.} {\sc and} {\sc Faber, W.} 2019.
\newblock Chain answer sets for logic programs with generalized atoms.
\newblock In {\em Proceedings of the Sixteenth European Conference on Logics in
  Artificial Intelligence (JELIA'19)}, {F.~Calimeri}, {N.~Leone}, {and}
  {M.~Manna}, Eds. Lecture Notes in Computer Science, vol. 11468.
  Springer-Verlag, 462--478.

\bibitem[\protect\citeauthoryear{Alviano, Faber, and Gebser}{Alviano
  et~al\mbox{.}}{2015}]{alfage15a}
{\sc Alviano, M.}, {\sc Faber, W.}, {\sc and} {\sc Gebser, M.} 2015.
\newblock Rewriting recursive aggregates in answer set programming: Back to
  monotonicity.
\newblock {\em Theory and Practice of Logic Programming\/}~{\em 15,\/}~4-5,
  559--573.

\bibitem[\protect\citeauthoryear{Alviano and Leone}{Alviano and
  Leone}{2015}]{alvleo15a}
{\sc Alviano, M.} {\sc and} {\sc Leone, N.} 2015.
\newblock Complexity and compilation of {GZ}-aggregates in answer set
  programming.
\newblock {\em Theory and Practice of Logic Programming\/}~{\em 15,\/}~4-5,
  574--587.

\bibitem[\protect\citeauthoryear{Apt, Blair, and Walker}{Apt
  et~al\mbox{.}}{1987}]{apblwa87a}
{\sc Apt, K.}, {\sc Blair, H.}, {\sc and} {\sc Walker, A.} 1987.
\newblock Towards a theory of declarative knowledge.
\newblock In {\em Foundations of Deductive Databases and Logic Programming},
  {J.~Minker}, Ed. Morgan Kaufmann Publishers, Chapter~2, 89--148.

\bibitem[\protect\citeauthoryear{Bailleux, Boufkhad, and Roussel}{Bailleux
  et~al\mbox{.}}{2009}]{baboro09a}
{\sc Bailleux, O.}, {\sc Boufkhad, Y.}, {\sc and} {\sc Roussel, O.} 2009.
\newblock New encodings of pseudo-{B}oolean constraints into {CNF}.
\newblock In {\em Proceedings of the Twelfth International Conference on Theory
  and Applications of Satisfiability Testing (SAT'09)}, {O.~Kullmann}, Ed.
  Lecture Notes in Computer Science, vol. 5584. Springer-Verlag, 181--194.

\bibitem[\protect\citeauthoryear{Bartholomew, Lee, and Meng}{Bartholomew
  et~al\mbox{.}}{2011}]{baleme11a}
{\sc Bartholomew, M.}, {\sc Lee, J.}, {\sc and} {\sc Meng, Y.} 2011.
\newblock First-order semantics of aggregates in answer set programming via
  modified circumscription.
\newblock In {\em Proceedings of the AAAI Spring Symposium on Logical
  Formalizations of Commonsense Reasoning}, {E.~Davis}, {P.~Doherty}, {and}
  {E.~Erdem}, Eds. AAAI Press, 16--22.

\bibitem[\protect\citeauthoryear{Ben-Eliyahu and Dechter}{Ben-Eliyahu and
  Dechter}{1994}]{bendec94a}
{\sc Ben-Eliyahu, R.} {\sc and} {\sc Dechter, R.} 1994.
\newblock Propositional semantics for disjunctive logic programs.
\newblock {\em Annals of Mathematics and Artificial Intelligence\/}~{\em
  12,\/}~1-2, 53--87.

\bibitem[\protect\citeauthoryear{Berman, Karpinski, Larmore, Plandowski, and
  Rytter}{Berman et~al\mbox{.}}{2002}]{bekalaplry02a}
{\sc Berman, P.}, {\sc Karpinski, M.}, {\sc Larmore, L.}, {\sc Plandowski, W.},
  {\sc and} {\sc Rytter, W.} 2002.
\newblock On the complexity of pattern matching for highly compressed
  two-dimensional texts.
\newblock {\em Journal of Computer and System Sciences\/}~{\em 65,\/}~2,
  332--350.

\bibitem[\protect\citeauthoryear{Bomanson, Gebser, and Janhunen}{Bomanson
  et~al\mbox{.}}{2014}]{bogeja14a}
{\sc Bomanson, J.}, {\sc Gebser, M.}, {\sc and} {\sc Janhunen, T.} 2014.
\newblock Improving the normalization of weight rules in answer set programs.
\newblock In {\em Proceedings of the Fourteenth European Conference on Logics
  in Artificial Intelligence (JELIA'14)}, {E.~Ferm{\'e}} {and} {J.~Leite}, Eds.
  Lecture Notes in Artificial Intelligence, vol. 8761. Springer-Verlag,
  166--180.

\bibitem[\protect\citeauthoryear{Bomanson, Janhunen, and Niemel{\"a}}{Bomanson
  et~al\mbox{.}}{2020}]{bojani20a}
{\sc Bomanson, J.}, {\sc Janhunen, T.}, {\sc and} {\sc Niemel{\"a}, I.} 2020.
\newblock Applying visible strong equivalence in answer-set program
  transformations.
\newblock {\em ACM Transactions on Computational Logic\/}~{\em 4,\/}~21,
  33:1--33:41.

\bibitem[\protect\citeauthoryear{Bomanson, Janhunen, and Weinzierl}{Bomanson
  et~al\mbox{.}}{2019}]{bojawe19a}
{\sc Bomanson, J.}, {\sc Janhunen, T.}, {\sc and} {\sc Weinzierl, A.} 2019.
\newblock Enhancing lazy grounding with lazy normalization in answer-set
  programming.
\newblock In {\em Proceedings of the Thirty-third National Conference on
  Artificial Intelligence (AAAI'19)}, {P.~{Van Hentenryck}} {and} {Z.~Zhou},
  Eds. AAAI Press, 2694--2702.

\bibitem[\protect\citeauthoryear{Brewka, Eiter, and Truszczy{\'n}ski}{Brewka
  et~al\mbox{.}}{2011}]{breitr11a}
{\sc Brewka, G.}, {\sc Eiter, T.}, {\sc and} {\sc Truszczy{\'n}ski, M.} 2011.
\newblock Answer set programming at a glance.
\newblock {\em Communications of the {ACM}\/}~{\em 54,\/}~12, 92--103.

\bibitem[\protect\citeauthoryear{Bruynooghe, Blockeel, Bogaerts, {De Cat}, {De
  Pooter}, Jansen, Labarre, Ramon, Denecker, and Verwer}{Bruynooghe
  et~al\mbox{.}}{2015}]{brblbocapojalaradeve15a}
{\sc Bruynooghe, M.}, {\sc Blockeel, H.}, {\sc Bogaerts, B.}, {\sc {De Cat},
  B.}, {\sc {De Pooter}, S.}, {\sc Jansen, J.}, {\sc Labarre, A.}, {\sc Ramon,
  J.}, {\sc Denecker, M.}, {\sc and} {\sc Verwer, S.} 2015.
\newblock Predicate logic as a modeling language: Modeling and solving some
  machine learning and data mining problems with {IDP}3.
\newblock {\em Theory and Practice of Logic Programming\/}~{\em 15,\/}~6,
  783--817.

\bibitem[\protect\citeauthoryear{Calimeri, Dodaro, Fusc{\`a}, Perri, and
  Zangari}{Calimeri et~al\mbox{.}}{2020}]{cadofupeza20a}
{\sc Calimeri, F.}, {\sc Dodaro, C.}, {\sc Fusc{\`a}, D.}, {\sc Perri, S.},
  {\sc and} {\sc Zangari, J.} 2020.
\newblock Efficiently coupling the {I-DLV} grounder with {ASP} solvers.
\newblock {\em Theory and Practice of Logic Programming\/}~{\em 20,\/}~2,
  205--224.

\bibitem[\protect\citeauthoryear{Calimeri, Faber, Gebser, Ianni, Kaminski,
  Krennwallner, Leone, Maratea, Ricca, and Schaub}{Calimeri
  et~al\mbox{.}}{2019}]{cafageiakakrlemarisc19a}
{\sc Calimeri, F.}, {\sc Faber, W.}, {\sc Gebser, M.}, {\sc Ianni, G.}, {\sc
  Kaminski, R.}, {\sc Krennwallner, T.}, {\sc Leone, N.}, {\sc Maratea, M.},
  {\sc Ricca, F.}, {\sc and} {\sc Schaub, T.} 2019.
\newblock {ASP-Core-2} input language format.
\newblock {\em Theory and Practice of Logic Programming\/}~{\em 20,\/}~2,
  294--309.

\bibitem[\protect\citeauthoryear{Codd}{Codd}{1970}]{codd70a}
{\sc Codd, E.} 1970.
\newblock A relational model of data for large shared data banks.
\newblock {\em Communications of the {ACM}\/}~{\em 13,\/}~6, 377--387.

\bibitem[\protect\citeauthoryear{Codd}{Codd}{1972}]{codd72a}
{\sc Codd, E.} 1972.
\newblock Relational completeness of data base sublanguages.
\newblock {\em Research Report / RJ / IBM / San Jose, California\/}~{\em
  RJ987}.

\bibitem[\protect\citeauthoryear{Cuteri, Dodaro, Ricca, and
  Sch{\"u}ller}{Cuteri et~al\mbox{.}}{2020}]{cudorisc20a}
{\sc Cuteri, B.}, {\sc Dodaro, C.}, {\sc Ricca, F.}, {\sc and} {\sc
  Sch{\"u}ller, P.} 2020.
\newblock Overcoming the grounding bottleneck due to constraints in {ASP}
  solving: Constraints become propagators.
\newblock In {\em Proceedings of the Twenty-Ninth International Joint
  Conference on Artificial Intelligence (IJCAI'20)}, {C.~Bessiere}, Ed.
  ijcai.org, 1688--1694.

\bibitem[\protect\citeauthoryear{Dantsin, Eiter, Gottlob, and Voronkov}{Dantsin
  et~al\mbox{.}}{2001}]{daeigovo01a}
{\sc Dantsin, E.}, {\sc Eiter, T.}, {\sc Gottlob, G.}, {\sc and} {\sc Voronkov,
  A.} 2001.
\newblock Complexity and expressive power of logic programming.
\newblock {\em {ACM} Computing Surveys\/}~{\em 33,\/}~3, 374--425.

\bibitem[\protect\citeauthoryear{Denecker, Marek, and
  Truszczy{\'n}ski}{Denecker et~al\mbox{.}}{2004}]{dematr04a}
{\sc Denecker, M.}, {\sc Marek, V.}, {\sc and} {\sc Truszczy{\'n}ski, M.} 2004.
\newblock Ultimate approximation and its application in nonmonotonic knowledge
  representation systems.
\newblock {\em Information and Computation\/}~{\em 192,\/}~1, 84--121.

\bibitem[\protect\citeauthoryear{Denecker, Pelov, and Bruynooghe}{Denecker
  et~al\mbox{.}}{2001}]{depebr01a}
{\sc Denecker, M.}, {\sc Pelov, N.}, {\sc and} {\sc Bruynooghe, M.} 2001.
\newblock Ultimate well-founded and stable semantics for logic programs with
  aggregates.
\newblock In {\em Proceedings of the Seventeenth International Conference on
  Logic Programming (ICLP'01)}, {P.~Codognet}, Ed. Lecture Notes in Computer
  Science, vol. 2237. Springer-Verlag, 212--226.

\bibitem[\protect\citeauthoryear{Eiter and Gottlob}{Eiter and
  Gottlob}{1995}]{eitgot95a}
{\sc Eiter, T.} {\sc and} {\sc Gottlob, G.} 1995.
\newblock On the computational cost of disjunctive logic programming:
  Propositional case.
\newblock {\em Annals of Mathematics and Artificial Intelligence\/}~{\em
  15,\/}~3-4, 289--323.

\bibitem[\protect\citeauthoryear{Eiter, Kaminski, Redl, and Weinzierl}{Eiter
  et~al\mbox{.}}{2018}]{eikarewe18a}
{\sc Eiter, T.}, {\sc Kaminski, T.}, {\sc Redl, C.}, {\sc and} {\sc Weinzierl,
  A.} 2018.
\newblock Exploiting partial assignments for efficient evaluation of answer set
  programs with external source access.
\newblock {\em Journal of Artificial Intelligence Research\/}~{\em 62},
  665--727.

\bibitem[\protect\citeauthoryear{Faber, Pfeifer, and Leone}{Faber
  et~al\mbox{.}}{2011}]{fapfle11a}
{\sc Faber, W.}, {\sc Pfeifer, G.}, {\sc and} {\sc Leone, N.} 2011.
\newblock Semantics and complexity of recursive aggregates in answer set
  programming.
\newblock {\em Artificial Intelligence\/}~{\em 175,\/}~1, 278--298.

\bibitem[\protect\citeauthoryear{Faber, Pfeifer, Leone, Dell'Armi, and
  Ielpa}{Faber et~al\mbox{.}}{2008}]{fapfledeie08a}
{\sc Faber, W.}, {\sc Pfeifer, G.}, {\sc Leone, N.}, {\sc Dell'Armi, T.}, {\sc
  and} {\sc Ielpa, G.} 2008.
\newblock Design and implementation of aggregate functions in the {DLV} system.
\newblock {\em Theory and Practice of Logic Programming\/}~{\em 8,\/}~5-6,
  545--580.

\bibitem[\protect\citeauthoryear{Ferraris}{Ferraris}{2011}]{ferraris11a}
{\sc Ferraris, P.} 2011.
\newblock Logic programs with propositional connectives and aggregates.
\newblock {\em ACM Transactions on Computational Logic\/}~{\em 12,\/}~4,
  25:1--25:40.

\bibitem[\protect\citeauthoryear{Ferraris and Lifschitz}{Ferraris and
  Lifschitz}{2005}]{ferlif02a}
{\sc Ferraris, P.} {\sc and} {\sc Lifschitz, V.} 2005.
\newblock Weight constraints as nested expressions.
\newblock {\em Theory and Practice of Logic Programming\/}~{\em 5,\/}~1-2,
  45--74.

\bibitem[\protect\citeauthoryear{Ganguly, Greco, and Zaniolo}{Ganguly
  et~al\mbox{.}}{1995}]{gagrza95a}
{\sc Ganguly, S.}, {\sc Greco, S.}, {\sc and} {\sc Zaniolo, C.} 1995.
\newblock Extrema predicates in deductive databases.
\newblock {\em Journal of Computer and System Sciences\/}~{\em 51,\/}~2,
  244--259.

\bibitem[\protect\citeauthoryear{Garey and Johnson}{Garey and
  Johnson}{1979}]{garjoh79}
{\sc Garey, M.} {\sc and} {\sc Johnson, D.} 1979.
\newblock {\em Computers and Intractability: A Guide to the Theory of
  {NP}-Completeness}.
\newblock W. Freeman and Co., New York.

\bibitem[\protect\citeauthoryear{Gebser, Harrison, Kaminski, Lifschitz, and
  Schaub}{Gebser et~al\mbox{.}}{2015a}]{gehakalisc15a}
{\sc Gebser, M.}, {\sc Harrison, A.}, {\sc Kaminski, R.}, {\sc Lifschitz, V.},
  {\sc and} {\sc Schaub, T.} 2015a.
\newblock Abstract {G}ringo.
\newblock {\em Theory and Practice of Logic Programming\/}~{\em 15,\/}~4-5,
  449--463.

\bibitem[\protect\citeauthoryear{Gebser, Janhunen, and Rintanen}{Gebser
  et~al\mbox{.}}{2014}]{gejari14c}
{\sc Gebser, M.}, {\sc Janhunen, T.}, {\sc and} {\sc Rintanen, J.} 2014.
\newblock Answer set programming as {SAT} modulo acyclicity.
\newblock In {\em Proceedings of the Twenty-first European Conference on
  Artificial Intelligence (ECAI'14)}, {T.~Schaub}, {G.~Friedrich}, {and}
  {B.~{O'Sullivan}}, Eds. IOS Press, 351--356.

\bibitem[\protect\citeauthoryear{Gebser, Kaminski, Kaufmann, and Schaub}{Gebser
  et~al\mbox{.}}{2009}]{gekakasc09a}
{\sc Gebser, M.}, {\sc Kaminski, R.}, {\sc Kaufmann, B.}, {\sc and} {\sc
  Schaub, T.} 2009.
\newblock On the implementation of weight constraint rules in conflict-driven
  {ASP} solvers.
\newblock In {\em Proceedings of the Twenty-fifth International Conference on
  Logic Programming (ICLP'09)}, {P.~Hill} {and} {D.~Warren}, Eds. Lecture Notes
  in Computer Science, vol. 5649. Springer-Verlag, 250--264.

\bibitem[\protect\citeauthoryear{Gebser, Kaminski, Kaufmann, and Schaub}{Gebser
  et~al\mbox{.}}{2019}]{gekakasc17a}
{\sc Gebser, M.}, {\sc Kaminski, R.}, {\sc Kaufmann, B.}, {\sc and} {\sc
  Schaub, T.} 2019.
\newblock Multi-shot {ASP} solving with clingo.
\newblock {\em Theory and Practice of Logic Programming\/}~{\em 19,\/}~1,
  27--82.

\bibitem[\protect\citeauthoryear{Gebser, Kaminski, and Schaub}{Gebser
  et~al\mbox{.}}{2015b}]{gekasc15a}
{\sc Gebser, M.}, {\sc Kaminski, R.}, {\sc and} {\sc Schaub, T.} 2015b.
\newblock Grounding recursive aggregates: Preliminary report.
\newblock In {\em Proceedings of the Third Workshop on Grounding, Transforming,
  and Modularizing Theories with Variables (GTTV'15)}, {M.~Denecker} {and}
  {T.~Janhunen}, Eds.

\bibitem[\protect\citeauthoryear{Gebser, Kaufmann, and Schaub}{Gebser
  et~al\mbox{.}}{2012}]{gekasc09c}
{\sc Gebser, M.}, {\sc Kaufmann, B.}, {\sc and} {\sc Schaub, T.} 2012.
\newblock Conflict-driven answer set solving: From theory to practice.
\newblock {\em Artificial Intelligence\/}~{\em 187-188}, 52--89.

\bibitem[\protect\citeauthoryear{Gelfond and Leone}{Gelfond and
  Leone}{2002}]{gelleo02a}
{\sc Gelfond, M.} {\sc and} {\sc Leone, N.} 2002.
\newblock Logic programming and knowledge representation --- the {A}-{P}rolog
  perspective.
\newblock {\em Artificial Intelligence\/}~{\em 138,\/}~1-2, 3--38.

\bibitem[\protect\citeauthoryear{Gelfond and Lifschitz}{Gelfond and
  Lifschitz}{1988}]{gellif88b}
{\sc Gelfond, M.} {\sc and} {\sc Lifschitz, V.} 1988.
\newblock The stable model semantics for logic programming.
\newblock In {\em Proceedings of the Fifth International Conference and
  Symposium of Logic Programming (ICLP'88)}, {R.~Kowalski} {and} {K.~Bowen},
  Eds. MIT Press, 1070--1080.

\bibitem[\protect\citeauthoryear{Gelfond and Lifschitz}{Gelfond and
  Lifschitz}{1991}]{gellif91a}
{\sc Gelfond, M.} {\sc and} {\sc Lifschitz, V.} 1991.
\newblock Classical negation in logic programs and disjunctive databases.
\newblock {\em New Generation Computing\/}~{\em 9}, 365--385.

\bibitem[\protect\citeauthoryear{Gelfond and Zhang}{Gelfond and
  Zhang}{2019}]{gelzha19a}
{\sc Gelfond, M.} {\sc and} {\sc Zhang, Y.} 2019.
\newblock Vicious circle principle, aggregates, and formation of sets in {ASP}
  based languages.
\newblock {\em Artificial Intelligence\/}~{\em 275}, 28--77.

\bibitem[\protect\citeauthoryear{Harrison and Lifschitz}{Harrison and
  Lifschitz}{2019}]{harlif19a}
{\sc Harrison, A.} {\sc and} {\sc Lifschitz, V.} 2019.
\newblock Relating two dialects of answer set programming.
\newblock {\em Theory and Practice of Logic Programming\/}~{\em 19,\/}~5-6,
  1006--1020.

\bibitem[\protect\citeauthoryear{H{\"o}lldobler, Manthey, and
  Steinke}{H{\"o}lldobler et~al\mbox{.}}{2012}]{homast12a}
{\sc H{\"o}lldobler, S.}, {\sc Manthey, N.}, {\sc and} {\sc Steinke, P.} 2012.
\newblock A compact encoding of pseudo-{B}oolean constraints into {SAT}.
\newblock In {\em Proceedings of the Thirty-fifth Annual German Conference on
  Artificial Intelligence (KI'12)}, {B.~Glimm} {and} {A.~Kr{\"u}ger}, Eds.
  Lecture Notes in Computer Science, vol. 7526. Springer-Verlag, 107--118.

\bibitem[\protect\citeauthoryear{Janhunen, Kaminski, Ostrowski, Schaub,
  Schellhorn, and Wanko}{Janhunen et~al\mbox{.}}{2017}]{jakaosscscwa17a}
{\sc Janhunen, T.}, {\sc Kaminski, R.}, {\sc Ostrowski, M.}, {\sc Schaub, T.},
  {\sc Schellhorn, S.}, {\sc and} {\sc Wanko, P.} 2017.
\newblock Clingo goes linear constraints over reals and integers.
\newblock {\em Theory and Practice of Logic Programming\/}~{\em 17,\/}~5-6,
  872--888.

\bibitem[\protect\citeauthoryear{Janhunen and Niemel{\"a}}{Janhunen and
  Niemel{\"a}}{2011}]{jannie11a}
{\sc Janhunen, T.} {\sc and} {\sc Niemel{\"a}, I.} 2011.
\newblock Compact translations of non-disjunctive answer set programs to
  propositional clauses.
\newblock In {\em Logic Programming, Knowledge Representation, and Nonmonotonic
  Reasoning: Essays Dedicated to {M}ichael {G}elfond on the Occasion of his
  65th Birthday}, {M.~Balduccini} {and} {T.~Son}, Eds. Lecture Notes in
  Computer Science, vol. 6565. Springer-Verlag, 111--130.

\bibitem[\protect\citeauthoryear{Kemp and Stuckey}{Kemp and
  Stuckey}{1991}]{kemstu91a}
{\sc Kemp, D.} {\sc and} {\sc Stuckey, P.} 1991.
\newblock Semantics of logic programs with aggregates.
\newblock In {\em Proceedings of the 1991 International Symposium on Logic
  Programming (ISLP'91)}, {V.~Saraswat} {and} {K.~Ueda}, Eds. MIT Press,
  387--401.

\bibitem[\protect\citeauthoryear{Klug}{Klug}{1982}]{klug82a}
{\sc Klug, A.} 1982.
\newblock Equivalence of relational algebra and relational calculus query
  languages having aggregate functions.
\newblock {\em Journal of the ACM\/}~{\em 29,\/}~3, 699--717.

\bibitem[\protect\citeauthoryear{Lee and Meng}{Lee and Meng}{2009}]{leemen09a}
{\sc Lee, J.} {\sc and} {\sc Meng, Y.} 2009.
\newblock On reductive semantics of aggregates in answer set programming.
\newblock In {\em Proceedings of the Tenth International Conference on Logic
  Programming and Nonmonotonic Reasoning (LPNMR'09)}, {E.~Erdem}, {F.~Lin},
  {and} {T.~Schaub}, Eds. Lecture Notes in Artificial Intelligence, vol. 5753.
  Springer-Verlag, 182--195.

\bibitem[\protect\citeauthoryear{Lierler and Susman}{Lierler and
  Susman}{2017}]{liesus17a}
{\sc Lierler, Y.} {\sc and} {\sc Susman, B.} 2017.
\newblock On relation between constraint answer set programming and
  satisfiability modulo theories.
\newblock {\em Theory and Practice of Logic Programming\/}~{\em 17,\/}~4,
  559--590.

\bibitem[\protect\citeauthoryear{Lifschitz}{Lifschitz}{2002}]{lifschitz02a}
{\sc Lifschitz, V.} 2002.
\newblock Answer set programming and plan generation.
\newblock {\em Artificial Intelligence\/}~{\em 138,\/}~1-2, 39--54.

\bibitem[\protect\citeauthoryear{Lifschitz, Pearce, and Valverde}{Lifschitz
  et~al\mbox{.}}{2001}]{lipeva01a}
{\sc Lifschitz, V.}, {\sc Pearce, D.}, {\sc and} {\sc Valverde, A.} 2001.
\newblock Strongly equivalent logic programs.
\newblock {\em ACM Transactions on Computational Logic\/}~{\em 2,\/}~4,
  526--541.

\bibitem[\protect\citeauthoryear{Liu and You}{Liu and You}{2013}]{liuyou13a}
{\sc Liu, G.} {\sc and} {\sc You, J.} 2013.
\newblock Relating weight constraint and aggregate programs: Semantics and
  representation.
\newblock {\em Theory and Practice of Logic Programming\/}~{\em 13,\/}~1,
  1--31.

\bibitem[\protect\citeauthoryear{Liu, Pontelli, Son, and Truszczy{\'n}ski}{Liu
  et~al\mbox{.}}{2010}]{liposotr10a}
{\sc Liu, L.}, {\sc Pontelli, E.}, {\sc Son, T.}, {\sc and} {\sc
  Truszczy{\'n}ski, M.} 2010.
\newblock Logic programs with abstract constraint atoms: The role of
  computations.
\newblock {\em Artificial Intelligence\/}~{\em 174,\/}~3-4, 295--315.

\bibitem[\protect\citeauthoryear{Liu and Truszczy{\'n}ski}{Liu and
  Truszczy{\'n}ski}{2006}]{liutru06a}
{\sc Liu, L.} {\sc and} {\sc Truszczy{\'n}ski, M.} 2006.
\newblock Properties and applications of programs with monotone and convex
  constraints.
\newblock {\em Journal of Artificial Intelligence Research\/}~{\em 27},
  299--334.

\bibitem[\protect\citeauthoryear{Lloyd}{Lloyd}{1987}]{lloyd87}
{\sc Lloyd, J.} 1987.
\newblock {\em Foundations of Logic Programming}.
\newblock 
Springer-Verlag.

\bibitem[\protect\citeauthoryear{Marek and Remmel}{Marek and
  Remmel}{2004}]{marrem04a}
{\sc Marek, V.} {\sc and} {\sc Remmel, J.} 2004.
\newblock Set constraints in logic programming.
\newblock In {\em Proceedings of the Seventh International Conference on Logic
  Programming and Nonmonotonic Reasoning (LPNMR'04)}, {V.~Lifschitz} {and}
  {I.~Niemel{\"a}}, Eds. Lecture Notes in Artificial Intelligence, vol. 2923.
  Springer-Verlag, 167--179.

\bibitem[\protect\citeauthoryear{Marek and Truszczy{\'n}ski}{Marek and
  Truszczy{\'n}ski}{1991}]{martru91c}
{\sc Marek, V.} {\sc and} {\sc Truszczy{\'n}ski, M.} 1991.
\newblock Autoepistemic logic.
\newblock {\em Journal of the ACM\/}~{\em 38,\/}~3, 588--619.

\bibitem[\protect\citeauthoryear{Marek and Truszczy{\'n}ski}{Marek and
  Truszczy{\'n}ski}{1999}]{martru99a}
{\sc Marek, V.} {\sc and} {\sc Truszczy{\'n}ski, M.} 1999.
\newblock Stable models and an alternative logic programming paradigm.
\newblock In {\em The Logic Programming Paradigm: A 25-Year Perspective},
  {K.~Apt}, {V.~Marek}, {M.~Truszczy{\'n}ski}, {and} {D.~Warren}, Eds.
  Springer-Verlag, 375--398.

\bibitem[\protect\citeauthoryear{Mazuran, Serra, and Zaniolo}{Mazuran
  et~al\mbox{.}}{2013}]{maseza13a}
{\sc Mazuran, M.}, {\sc Serra, E.}, {\sc and} {\sc Zaniolo, C.} 2013.
\newblock Extending the power of {D}atalog recursion.
\newblock {\em Journal on Very Large Data Bases\/}~{\em 22,\/}~4, 471--493.

\bibitem[\protect\citeauthoryear{Mumick, Pirahesh, and Ramakrishnan}{Mumick
  et~al\mbox{.}}{1990}]{mupira90a}
{\sc Mumick, I.}, {\sc Pirahesh, H.}, {\sc and} {\sc Ramakrishnan, R.} 1990.
\newblock The magic of duplicates and aggregates.
\newblock In {\em Proceedings of the Sixteenth International Conference on Very
  Large Data Bases (VLDB'90)}, {D.~McLeod}, {R.~Sacks{-}Davis}, {and}
  {H.~Schek}, Eds. Morgan Kaufmann Publishers, 264--277.

\bibitem[\protect\citeauthoryear{Niemel{\"a}}{Niemel{\"a}}{1999}]{niemela99a}
{\sc Niemel{\"a}, I.} 1999.
\newblock Logic programs with stable model semantics as a constraint
  programming paradigm.
\newblock {\em Annals of Mathematics and Artificial Intelligence\/}~{\em
  25,\/}~3-4, 241--273.

\bibitem[\protect\citeauthoryear{{\"O}zsoyoglu, {\"O}zsoyoglu, and
  Matos}{{\"O}zsoyoglu et~al\mbox{.}}{1987}]{ozozma87a}
{\sc {\"O}zsoyoglu, G.}, {\sc {\"O}zsoyoglu, Z.}, {\sc and} {\sc Matos, V.}
  1987.
\newblock Extending relational algebra and relational calculus with set-valued
  attributes and aggregate functions.
\newblock {\em ACM Transactions on Database Systems\/}~{\em 12,\/}~4, 566--592.

\bibitem[\protect\citeauthoryear{Pelov, Denecker, and Bruynooghe}{Pelov
  et~al\mbox{.}}{2003}]{pedebr03a}
{\sc Pelov, N.}, {\sc Denecker, M.}, {\sc and} {\sc Bruynooghe, M.} 2003.
\newblock Translation of aggregate programs to normal logic programs.
\newblock In {\em Proceedings of the Second International Workshop on Answer
  Set Programming (ASP'03)}, {M.~{de Vos}} {and} {A.~Provetti}, Eds. 
  CEUR Workshop Proceedings (CEUR-WS.org), 29--42.

\bibitem[\protect\citeauthoryear{Pelov, Denecker, and Bruynooghe}{Pelov
  et~al\mbox{.}}{2007}]{pedebr07a}
{\sc Pelov, N.}, {\sc Denecker, M.}, {\sc and} {\sc Bruynooghe, M.} 2007.
\newblock Well-founded and stable semantics of logic programs with aggregates.
\newblock {\em Theory and Practice of Logic Programming\/}~{\em 7,\/}~3,
  301--353.

\bibitem[\protect\citeauthoryear{Reiter}{Reiter}{1977}]{reiter77a}
{\sc Reiter, R.} 1977.
\newblock On closed world data bases.
\newblock In {\em Proceedings of Workshop on Logic and Databases},
  {H.~Gallaire} {and} {J.~Minker}, Eds. Plenum Press, 119--140.

\bibitem[\protect\citeauthoryear{Ross}{Ross}{1994}]{ross94a}
{\sc Ross, K.} 1994.
\newblock Modular stratification and magic sets for {D}atalog programs with
  negation.
\newblock {\em Journal of the ACM\/}~{\em 41,\/}~6, 1216--1266.

\bibitem[\protect\citeauthoryear{Roussel and Manquinho}{Roussel and
  Manquinho}{2009}]{rouman09a}
{\sc Roussel, O.} {\sc and} {\sc Manquinho, V.} 2009.
\newblock Pseudo-{B}oolean and cardinality constraints.
\newblock In {\em Handbook of Satisfiability}, {A.~Biere}, {M.~Heule}, {H.~{van
  Maaren}}, {and} {T.~Walsh}, Eds. 
IOS Press, Chapter~22, 695--733.

\bibitem[\protect\citeauthoryear{Schlipf}{Schlipf}{1995}]{schlipf95a}
{\sc Schlipf, J.} 1995.
\newblock The expressive powers of the logic programming semantics.
\newblock {\em Journal of Computer and System Sciences\/}~{\em 51}, 64--86.

\bibitem[\protect\citeauthoryear{Shen, Wang, Eiter, Fink, Redl, Krennwallner,
  and Deng}{Shen et~al\mbox{.}}{2014}]{shwaeifirekrde14a}
{\sc Shen, Y.}, {\sc Wang, K.}, {\sc Eiter, T.}, {\sc Fink, M.}, {\sc Redl,
  C.}, {\sc Krennwallner, T.}, {\sc and} {\sc Deng, J.} 2014.
\newblock {FLP} answer set semantics without circular justifications for
  general logic programs.
\newblock {\em Artificial Intelligence\/}~{\em 213}, 1--41.

\bibitem[\protect\citeauthoryear{Simons, Niemel{\"a}, and Soininen}{Simons
  et~al\mbox{.}}{2002}]{siniso02a}
{\sc Simons, P.}, {\sc Niemel{\"a}, I.}, {\sc and} {\sc Soininen, T.} 2002.
\newblock Extending and implementing the stable model semantics.
\newblock {\em Artificial Intelligence\/}~{\em 138,\/}~1-2, 181--234.

\bibitem[\protect\citeauthoryear{Son and Pontelli}{Son and
  Pontelli}{2007}]{sonpon07b}
{\sc Son, T.} {\sc and} {\sc Pontelli, E.} 2007.
\newblock A constructive semantic characterization of aggregates in answer set
  programming.
\newblock {\em Theory and Practice of Logic Programming\/}~{\em 7,\/}~3,
  355--375.

\bibitem[\protect\citeauthoryear{Son, Pontelli, and Elkabani}{Son
  et~al\mbox{.}}{2006}]{sopoel06a}
{\sc Son, T.}, {\sc Pontelli, E.}, {\sc and} {\sc Elkabani, I.} 2006.
\newblock An unfolding-based semantics for logic programming with aggregates.
\newblock {\em CoRR\/}~{\em abs/cs/0605038}.

\bibitem[\protect\citeauthoryear{Sudarshan and Ramakrishnan}{Sudarshan and
  Ramakrishnan}{1991}]{sudram91a}
{\sc Sudarshan, S.} {\sc and} {\sc Ramakrishnan, R.} 1991.
\newblock Aggregation and relevance in deductive databases.
\newblock In {\em Proceedings of the Seventeenth International Conference on
  Very Large Data Bases (VLDB'91)}, {G.~Lohman}, {A.~Sernadas}, {and}
  {R.~Camps}, Eds. Morgan Kaufmann Publishers, 501--511.

\bibitem[\protect\citeauthoryear{Syrj{\"a}nen}{Syrj{\"a}nen}{2001}]{lparseManual}
{\sc Syrj{\"a}nen, T.} 2001.
\newblock Lparse 1.0 user's manual.
\newblock www.tcs.hut.fi/Software/smodels/.

\bibitem[\protect\citeauthoryear{Truszczy{\'n}ski}{Truszczy{\'n}ski}{2010}]{truszczynski10a}
{\sc Truszczy{\'n}ski, M.} 2010.
\newblock Reducts of propositional theories, satisfiability relations, and
  generalizations of semantics of logic programs.
\newblock {\em Artificial Intelligence\/}~{\em 174,\/}~16-17, 1285--1306.

\bibitem[\protect\citeauthoryear{{van Emden} and Kowalski}{{van Emden} and
  Kowalski}{1976}]{emdkow76a}
{\sc {van Emden}, M.} {\sc and} {\sc Kowalski, R.} 1976.
\newblock The semantics of predicate logic as a programming language.
\newblock {\em Journal of the ACM\/}~{\em 23,\/}~4, 733--742.

\bibitem[\protect\citeauthoryear{{Van Gelder}}{{Van Gelder}}{1992}]{gelder92a}
{\sc {Van Gelder}, A.} 1992.
\newblock The well-founded semantics of aggregation.
\newblock In {\em Proceedings of the Eleventh ACM SIGACT-SIGMOD-SIGART
  Symposium on Principles of Database Systems (PODS'92)}, {M.~Vardi} {and}
  {P.~Kanellakis}, Eds. ACM Press, 127--138.

\bibitem[\protect\citeauthoryear{Vanbesien, Bruynooghe, and Denecker}{Vanbesien
  et~al\mbox{.}}{2021}]{DBLP:journals/corr/abs-2104-14789}
{\sc Vanbesien, L.}, {\sc Bruynooghe, M.}, {\sc and} {\sc Denecker, M.} 2021.
\newblock Analyzing semantics of aggregate answer set programming using
  approximation fixpoint theory.
\newblock {\em CoRR\/}~{\em abs/2104.14789}.

\bibitem[\protect\citeauthoryear{Weinzierl, Taupe, and Friedrich}{Weinzierl
  et~al\mbox{.}}{2020}]{wetafr20a}
{\sc Weinzierl, A.}, {\sc Taupe, R.}, {\sc and} {\sc Friedrich, G.} 2020.
\newblock Advancing lazy-grounding {ASP} solving techniques --- restarts, phase
  saving, heuristics, and more.
\newblock {\em Theory and Practice of Logic Programming\/}~{\em 20,\/}~5,
  609--624.

\bibitem[\protect\citeauthoryear{Zaniolo, Yang, Das, Shkapsky, Condie, and
  Interlandi}{Zaniolo et~al\mbox{.}}{2017}]{zayadashcoin17a}
{\sc Zaniolo, C.}, {\sc Yang, M.}, {\sc Das, A.}, {\sc Shkapsky, A.}, {\sc
  Condie, T.}, {\sc and} {\sc Interlandi, M.} 2017.
\newblock Fixpoint semantics and optimization of recursive {D}atalog programs
  with aggregates.
\newblock {\em Theory and Practice of Logic Programming\/}~{\em 17,\/}~5-6,
  1048--1065.

\end{thebibliography}

\end{document}
